%% file: gnf_main.tex
\PassOptionsToPackage{numbers}{natbib}
\documentclass{article}

\usepackage[preprint]{neurips_2019}

\usepackage[utf8]{inputenc} 
\usepackage[T1]{fontenc}    
\usepackage{hyperref}       
\usepackage{url}            
\usepackage{booktabs}       
\usepackage{amsfonts}       
\usepackage{nicefrac}       
\usepackage{microtype}      
\usepackage{graphicx}
\usepackage{caption}
\usepackage{subcaption}

\usepackage{amsmath}
\usepackage{amssymb}
\usepackage[dvipsnames]{xcolor}
\usepackage{todonotes}
\usepackage{wrapfig,lipsum,booktabs}
\usepackage{nicefrac}

\input{notation}

\title{Graph Normalizing Flows}

\newcommand*\samethanks[1][\value{footnote}]{\footnotemark[#1]}

\author{Jenny Liu\thanks{Equal Contribution} \\
    University of Toronto \\
    Vector Institute \\
    \texttt{jyliu@cs.toronto.edu}
    \And
    Aviral Kumar\samethanks\ \,\thanks{Work done during an internship at Google} \\
    UC Berkeley \\
    \texttt{aviralk@berkeley.edu}
    \And
    Jimmy Ba \\
    University of Toronto \\
    Vector Institute \\
    \texttt{jba@cs.toronto.edu}
    \And
    Jamie Kiros \\
    Google Research \\
    \texttt{kiros@google.com}
    \And
    Kevin Swersky \\
    Google Research \\
    \texttt{kswersky@google.com}
}
\begin{document}
\linespread{0.99}\selectfont

\maketitle

\begin{abstract}
  We introduce graph normalizing flows: a new, reversible graph neural network model for prediction and generation. On supervised tasks, graph normalizing flows perform similarly to message passing neural networks, but at a significantly reduced memory footprint, allowing them to scale to larger graphs. In the unsupervised case, we combine graph normalizing flows with a novel graph auto-encoder to create a generative model of graph structures. Our model is permutation-invariant, generating entire graphs with a single feed-forward pass, and achieves competitive results with the state-of-the art auto-regressive models, while being better suited to parallel computing architectures.
\end{abstract}

\input{sections/introduction}
\input{sections/background}
\input{sections/methods}
\input{sections/experiments}
\input{sections/conclusion}
\small
\bibliography{gnf_main}
\bibliographystyle{plainnat}
\newpage
\input{sections/supplementary_material}


\end{document}

%% file: notation.tex
\newcommand{\neighborhood}{\mathcal{N}}
\newcommand{\agg}{\mathsf{Agg}}

\newcommand{\bx}{\mathbf{x}}
\newcommand{\bz}{\mathbf{z}}
\newcommand{\bh}{\mathbf{h}}
\newcommand{\bm}{\mathbf{m}}
\newcommand{\bv}{v}
\newcommand{\bu}{u}
\newcommand{\bw}{w}

\newcommand{\bxzero}{\bx^{(0)}}
\newcommand{\bxone}{\bx^{(1)}}
\newcommand{\bzzero}{\bz^{(0)}}
\newcommand{\bzone}{\bz^{(1)}}

\newcommand{\params}{\boldsymbol{\theta}}

\newcommand{\community}{\textsc{community-small}}
\newcommand{\ego}{\textsc{ego-small}}
\newcommand{\graphvae}{\textsc{GraphVAE}}
\newcommand{\deepgmg}{\textsc{DeepGMG}}

\newcommand{\graphrnn}{\textsc{GraphRNN}}
\newcommand{\gnf}{\textsc{GNF}}
\newcommand{\realnvp}{\textsc{RealNVP}}

%% file: sections/introduction.tex
\section{Introduction}
\label{sec:introduction}
Graph-structured data is ubiquitous in science and engineering, and modeling graphs is an important component of being able to make predictions and reason about these domains. Machine learning has recently turned its attention to modeling graph-structured data using graph neural networks (GNNs)~\cite{gori2005new, scarselli2009graph, li2015gated, kipf2016semi, gilmer17_mpnn} that can exploit the structure of relational systems to create more accurate and generalizable predictions. For example, these can be used to predict the properties of molecules in order to aid in search and discovery~\cite{duvenaud2015convolutional, gilmer17_mpnn}, or to learn physical properties of robots such that new robots with different structures can be controlled without re-learning a control policy~\cite{wang2018nervenet}.

In this paper, we introduce a new formulation for graph neural networks by extending the framework of normalizing flows~\cite{rezende15normalizing, dinh2014nice, dinh2016density} to graph-structured data. We call these models graph normalizing flows (GNFs). GNFs have the property that the message passing computation is exactly reversible, meaning that one can exactly reconstruct the input node features from the GNN representation; this results in GNFs having several useful properties.

In the supervised case, we leverage a similar mechanism to~\cite{gomez17_revnet} to obtain significant memory savings in a model we call reversible graph neural networks, or GRevNets. Ordinary GNNs require the storage of hidden states after every message passing step in order to facilitate backpropagation. This means one needs to store $O(\text{\#nodes}\times\text{\#message passing steps})$ states, which can be costly for large graphs. In contrast, GRevNets can reconstruct hidden states in lower layers from higher layers during backpropagation, meaning one only needs to store $O(\text{\#nodes})$ states. A recent approach for memory saving based on recurrent backpropagation (RBP)~\cite{Almeida87_rbp, pineda87_rbp, liao18_rbp} requires running message passing to convergence, followed by the approximate, iterative inversion of a large matrix. Conversely, GRevNets get the exact gradients at a minor additional cost, equivalent to one extra forward pass. We show that GRevNets are competitive with conventional memory-inefficient GNNs, and outperform RBP on standard benchmarks.

In the unsupervised case, we use GNFs to develop a generative model of graphs. Learned generative models of graphs are a relatively new and less explored area. Machine learning has been quite successful at generative modeling of complex domains such as images, audio, and text. However, relational data poses new and interesting challenges such as permutation invariance, as permuting the nodes results in the same underlying graph.

One of the most successful approaches so far is to model the graph using an auto-regressive process~\cite{li2018deepgen,you18graphrnn}. These generate each node in sequence, and for each newly generated node, the corresponding edges to previously generated nodes are also created. In theory, this is capable of modeling the full joint distribution, but computing the full likelihood requires marginalizing over all possible node-orderings. Sequential generation using RNNs also potentially suffers from trying to model long-range dependencies.

Normalizing flows are primarily designed for continuous-valued data, and the GNF models a distribution over a structured, continuous space over sets of variables. We combine this with a novel permutation-invariant graph auto-encoder to generate embeddings that are decoded into an adjacency matrix in a similar manner to~\cite{li2016gmmn, makhzani2016adversarial}. The result is a fully permutation-invariant model that achieves competitive results compared to GraphRNN~\cite{you18graphrnn}, while being more well-suited to parallel computing architectures.

%% file: sections/background.tex
\section{Background}
\subsection{Graph Neural Networks}
\label{sec:background-gnns}
\textbf{Notation: }A graph is defined as $\mathcal{G} = (H, \Omega)$, where $H \in \mathbb{R}^{N \times d_n}, H = (\bh^{(1)}, \cdots, \bh^{(N)})$ is the node feature matrix consisting of node features, of size $d_n$, for each of the $N$ nodes ($\bh^{(\bv)}$ for node $\bv$) in the graph. $\Omega \in \mathbb{R}^{N \times N \times (d_e + 1)}$ is the edge feature matrix for the graph. The first channel of $\Omega$ is the \emph{adjacency matrix} of the graph (i.e. $\Omega_{i, j, 0} = 1$ if $e_{ij}$ is an edge in the graph). The rest of the matrix $\Omega_{i,j, 1:(d_e+1)}$ is the set of edge features of size $d_e$ for each possible edge $(i, j)$ in the graph. 

Graph Neural Networks (GNNs) or Message Passing Neural Nets (MPNNs)~\cite{gilmer17_mpnn} are a generalization/unification of a number of neural net architectures on graphs used in literature for a variety of tasks ranging from molecular modeling to network relational modeling. In general, MPNNs have two phases in the forward pass -- a message passing (MP) phase and a readout (R) phase.  The MP phase runs for $T$ time steps, $t=1,\ldots,T$ and is defined in terms of message generation functions $M_t$ and vertex update functions $U_t$. During each step in the message passing phase, hidden node features $\bh_t^{(\bv)}$ at each node in the graph are updated based on messages $\bm_{t+1}^{(\bv)}$ according to 
\begin{align}
\vspace{-20pt}
    \bm_{t+1}^{(\bv)} &=  \agg\left (\left \{M_t(\bh_t^{(\bv)}, \bh_t^{(\bu)}, \Omega_{\bu, \bv})\right \}_{\bu \in \neighborhood(\bv)}\right ) \label{eq:mp-messages} \\
    \bh_{t+1}^{(\bv)} &= U_t (\bh_t^{(\bv)}, \bm_{t+1}^{(\bv)}) \label{eq:mp-updatestate}
\end{align}
where $\agg$ is an aggregation function (e.g., sum), and $\neighborhood(\bv)$ denotes the set of neighbours to node $\bv$ in the graph. The R phase converts the final node embeddings at MP step $T$ into task-specific features by e.g., max-pooling.

One particularly useful aggregation function is graph attention~\cite{v2018_gat}, which uses attention~\cite{bahdanau2014neural, vaswani2017attention} to weight the messages from adjacent nodes. This involves computing an attention coefficient $\alpha$ between adjacent nodes using a linear transformation $W$, an attention mechanism $a$, and a nonlinearity $\sigma$,
\begin{align*}
    e_{t+1}^{(\bv, \bu)} &= a(W\bh_{t}^{(\bv)}, W\bh_{t}^{(\bu)}), \ \ \ \ \ \
    \alpha_{t+1}^{\bv, \bu} = \frac{\exp(e_{t+1}^{(\bv, \bu)})}{\sum_{\bw \in \neighborhood(\bv)} \exp(e_{t+1}^{(\bu, \bw)})} \\
    \bm_{t+1}^{(\bv)} &=  \sigma\Big(\sum_{\bu \in \neighborhood(\bv)} \alpha_{t+1}^{(\bv, \bu)} M(\bh_{t}^{(\bv)}, \bh_{t}^{(\bu)}, \Omega_{\bu, \bv}) \Big)
\end{align*}
Multi-headed attention~\cite{vaswani2017attention} applies attention with multiple weights $W$ and concatenates the results.

\subsection{Normalizing Flows}
Normalizing flows (NFs)~\cite{rezende15normalizing, dinh2014nice, dinh2016density} are a class of generative models that use invertible mappings to transform an observed vector $\bx \in \mathbb{R}^d$ to a latent vector $\bz \in \mathbb{R}^d$ using a mapping function $\bz = f(\bx)$ with inverse $\bx=f^{-1}(f(\bx))$. The change of variables formula relates a density function over $\bx$, $P(\bx)$ to one over $\bz$ by
\begin{align*}
    P(\bz) &= P(\bx)\bigg|\frac{\partial f(\bx)}{\partial \bx}\bigg|^{-1}
\end{align*}
With a sufficiently expressive mapping, NFs can learn to map a complicated distribution into one that is well modeled as a Gaussian; the key is to find a mapping that is expressive, but with an efficiently computable determinant. We base our formulation on non-volume preserving flows, a.k.a RealNVP~\cite{dinh2016density}. Specifically, the affine coupling layer involves partitioning the dimensions of $\bx$ into two sets of variables, $\bxzero$ and $\bxone$, and mapping them onto variables $\bzzero$ and $\bzone$ by
\begin{align*}
    \bzzero &= \bxzero \\
    \bzone &= \bxone \odot \exp(s(\bxzero)) + t(\bxzero)
\end{align*}
Where $s$ and $t$ are nonlinear functions and $\odot$ is the Hadamard product. The resulting Jacobian is lower triangular and its determinant is therefore efficiently computable.

%% file: sections/methods.tex
\section{Methods}
\subsection{Reversible Graph Neural Networks (GRevNets)}
GRevNets are a family of reversible message passing neural network models. To achieve reversibility, the node feature matrix of a GNN is split into two parts along the feature dimension--$H_t^{(0)}$ and $H_t^{(1)}$. For a particular node in the graph $\bv$, the two parts of its features at time $t$ in the message passing phase are called $\bh_t^0$ and $\bh_t^1$ respectively, such that $\bh_t^{(\bv)} = \mathsf{concat(}\bh_t^0, \bh_t^1\mathsf{)}$.

\begin{figure*}[h!]
    \centering
    \begin{subfigure}[t]{0.66\textwidth}
    \includegraphics[width=0.8\linewidth]{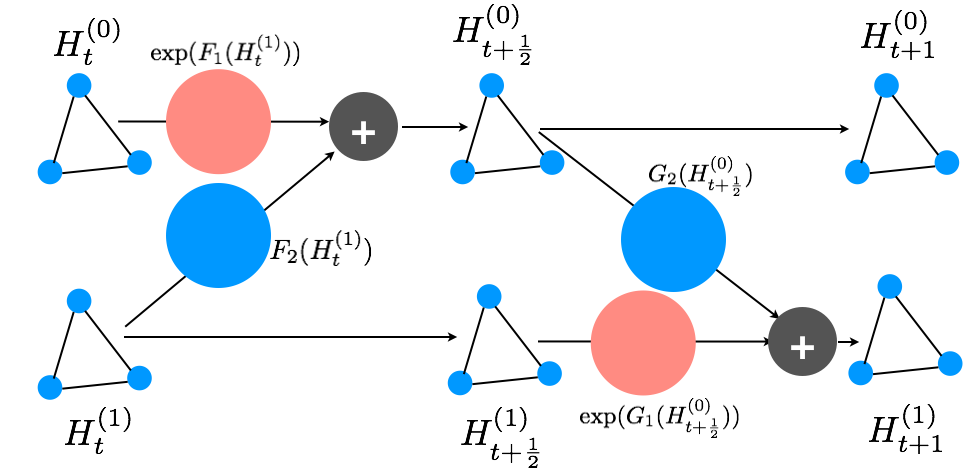}
    \caption{Architecture of 1 step of message passing in a GRevNet: $H_t^{(0)}$, $H_t^{(1)}$ denote the two parts of the node-features of a particular node, $F_1(\cdot), F_2(\cdot)$ and $G_1(\cdot), G_2(\cdot)$ are 1-step MP transforms consisting of applying $M_t$ and $U_t$ once each. The scaling functions ($F_2, G_2$) are shown in red, whereas the translation functions $(F_1, G_1)$ are shown in blue.}
    \label{fig:grevnet_fig}
    \end{subfigure}
    ~
    \begin{subfigure}[t]{0.31\textwidth}
    \includegraphics[width=0.99\linewidth]{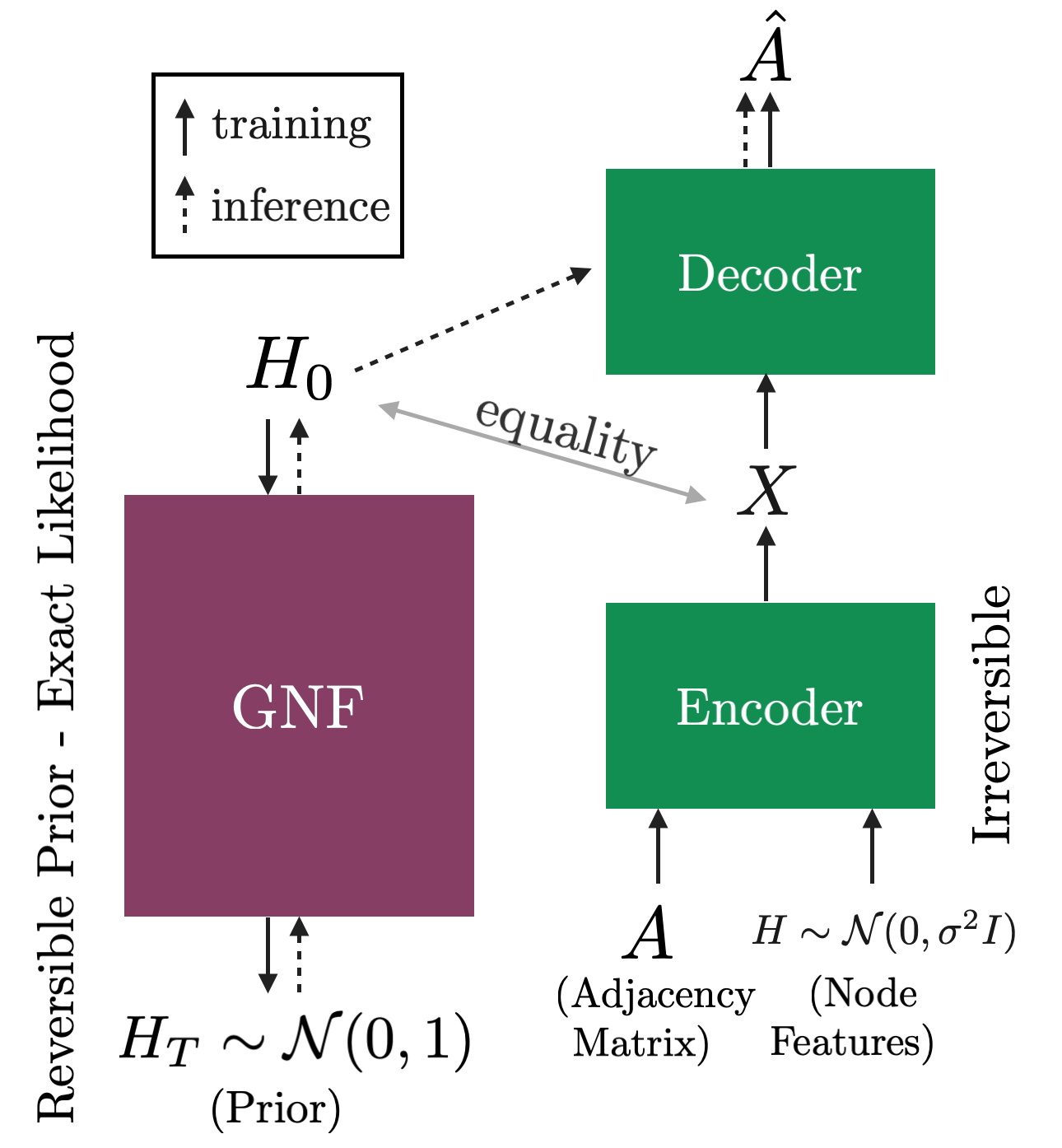}
    \caption{A summary of our Graph generation pipeline using GNFs. The GNF model generates node features which are then fed into a deocder to generate the adjacency.}
    \end{subfigure}
    \caption{(a) GRevnet Message Passing, and (b) GNF generation pipeline }
    \label{fig:graph_gen}
\end{figure*}

One step of the message passing procedure is broken down into into two intermediate steps, each of which is denoted as a half-step. $F_1(\cdot)$, $F_2(\cdot)$, $G_1(\cdot)$, and $G_2(\cdot)$ denote instances of the MP transformation given in Equations~\eqref{eq:mp-messages} and~\eqref{eq:mp-updatestate}, with $F_1/G_1$ and $F_2/G_2$ indicating whether the function is applied to scaling or translation. These functions consist of applying $M_t$ and then $U_t$ to one set of the partitioned features, given the graph adjacency matrix $\Omega$.  Figure~\ref{fig:grevnet_fig} depicts the procedure in detail.
\begin{align}
    H_{t+\frac{1}{2}}^{(0)} &= H_{t}^{(0)} \odot \exp\left(F_1\left(H_t^{(1)}\right)\right) + F_2(H_t^{(1)})
    \qquad\qquad H_{t+1}^{(0)} = H_{t+\frac{1}{2}}^{(0)} \nonumber \\
    H_{t+\frac{1}{2}}^{(1)} &= H_t^{(1)}
    \qquad\qquad\qquad\qquad\qquad H_{t+1}^{(1)} = H_{t+\frac{1}{2}}^{(1)} \odot \exp\left(G_1\left(H_{t+\frac{1}{2}}^{(0)}\right)\right) + G_2\left(H_{t+\frac{1}{2}}^{(0)}\right)
    \vspace{-0.1in}
    \label{eqn:reversible_eqns} &
    \vspace{-0.1in}
\end{align}
This architecture is easily reversible given $H_{t+1}^{(0)}$ and $H_{t+1}^{(1)}$, with the reverse procedure given by,
\begin{align}
    H_{t+\frac{1}{2}}^{(0)} &= H_{t+1}^{(0)} & H_t^{(1)} &= H_{t+\frac{1}{2}}^{(1)} \nonumber \\
    H_{t+\frac{1}{2}}^{(1)} &= \frac{\left(H_{t+1}^{(1)} - G_2\left (H_{t+\frac{1}{2}}^{(0)}\right)\right )}{\exp\left(G_1\left(H_{t+\frac{1}{2}}^{(0)}\right)\right)} & H_{t}^{(0)} &= \frac{\left(H_{t+\frac{1}{2}}^{(0)} - F_2(H_t^{(1)})\right)}{\exp\left(F_1\left(H_t^{(1)}\right)\right)}
    \vspace{-0.2in}
\end{align}

\subsection{GNFs for Structured Density Estimation}
In the same spirit as NFs, we can use the change of variables to give us the rule for exact density transformation. If we assume $H_t \sim P(H_t)$, then the density in terms of $P(H_{t-1})$ is given by $P(H_{t-1}) = \det \bigg|\frac{\partial H_{t-1}}{\partial H_t} \bigg| P(H_{t})$ so that
\begin{alignat*}{2}
    P(\mathcal{G}) &= \det \bigg|\frac{\partial H_{T}}{\partial H_0} \bigg| P(H_{T}) &= P(H_{T}) \prod_{t=1}^{T} \det \bigg|\frac{\partial H_{t}}{\partial H_{t-1}} \bigg|
\end{alignat*}
with $H_0$ being the input node features. The Jacobians are given by lower triangular matrices, hence making density computations tractable.

GNFs can model expressive distributions in continuous spaces over \emph{sets} of vectors. We choose the prior $P(H_T)=\prod_{i=1}^{N} \neighborhood(\bh_i | 0, I)$ to be a product of independent, standard Gaussian distributions. Sampling simply involves sampling a set of Gaussian vectors and running the inverse mapping. One free variable is the number of nodes that must be generated before initiating message passing. We simply model this as a fixed prior $P(N)$, where the distribution is given by the empirical distribution in the training set. Sampling graphs uniformly from the training set is equivalent to sampling $N$ from this distribution, and then sampling $\mathcal{G}$ uniformly from the set of training graphs with $N$ nodes.

Notice that the graph message passing induces dependencies between the nodes in the input space, which is reflected in the Jacobian. This also allows us to cast the RealNVP in the GNF framework: simply remove the edges from the graph so that all nodes become independent. Then each node transformation will be a sequence of reversible non-linear mappings, with no between-node dependencies\footnote{This formulation is specifically applicable to unstructured vector spaces, as opposed to images, which would involve checkerboard partitions and other domain-specific heuristics.}. We use this as a baseline to demonstrate that the GNF benefits when the nodes must model dependencies between each other.

In the absence of a known graph structure, as is the case for generation, we use a fully connected graph neural network. This allows the model to learn how to organize nodes in order to match a specific distribution. However, this poses a problem for certain aggregation functions like sum and mean. In the sum case, the message variance will increase with the number of nodes, and in both sum and mean cases, the messages from each node will have to contend with the messages from every other node. If there is a salient piece of information being sent from one node to another, then it could get drowned out by less informative messages. Instead, we opt to use graph attention as discussed in Section~\ref{sec:background-gnns}. This allows each node to choose the messages that it deems to be the most informative.

The result of using a fully connected graph is that the computational cost of message passing is $O(N^2)$, similar to the GraphRNN. However, each step of the GNF is expressible in terms of matrix operations, making it more amenable to parallel architectures. This is a similar justification for using transformers over RNNs~\cite{vaswani2017attention}.

\subsection{Graph Auto-Encoders}
\label{sec:graph-auto-encoders}
While GNFs are expressive models for structured, continuous spaces, our objective is to train a generative model of graph structures, an inherently discrete problem. Our strategy to solve this is to use a two-step process: (1) train a permutation invariant graph auto-encoder to create a graph encoder that embeds graphs into a continuous space; (2) train a GNF to model the distribution of the graph embeddings, and use the decoder to generate graphs. Each stage is trained separately. A similar strategy has been employed in prior work on generative models in~\cite{li2016gmmn, makhzani2016adversarial}. 

Note that in contrast to the GraphVAE~\cite{kipf16graphvae}, which generates a single vector to model the entire graph, we instead embed a set of nodes in a graph jointly, but each node will be mapped to its own embedding vector. This avoids the issue of having to run a matching process in the decoder.

The graph auto-encoder takes in a graph $\mathcal{G}$ and reconstructs the elements of the adjacency matrix, $A$, where $A_{ij}=1$ if node $\bv_i$ has an edge connecting it to node $\bv_j$, and $0$ otherwise. We focus on undirected graphs, meaning that we only need to predict the upper (or lower) triangular portion of $A$, but this methodology could easily extend to directed graphs.

The encoder takes in a set of node features $H \in \mathbb{R}^{N \times d}$ and an adjacency matrix $A \in \left \{ 0, 1\right \}^{N \times \frac{N}{2}}$ ($\frac{N}{2}$ since the graph is undirected) and outputs a set of node embeddings $X \in \mathbb{R}^{N \times k}$. The decoder takes these embeddings and outputs a set of edge probabilities $\hat{A} \in \left [0, 1 \right ]^{N \times \frac{N}{2}}$. For parameters $\params$, we use the binary cross entropy loss function,
\begin{equation}
    \mathcal{L}(\params) = -\sum_{i=1}^{N} \sum_{j=1}^{\frac{N}{2}} A_{ij} \log(\hat{A}_{ij}) + (1 - A_{ij})\log(1 - \hat{A}_{ij}). \label{eqn:graph_autoencoder_loss}
\end{equation}
We use a relatively simple decoder. Given node embeddings $\bx_{i}$ and $\bx_{j}$, our decoder outputs the edge probability as 
\begin{equation}
\hat{A}_{ij} = \frac{1}{1 + \exp(C(\|\bx_i - \bx_j\|_2^2 - 1))}
\end{equation}
where $C$ is a temperature hyperparameter, set to 10 in our experiments. This reflects the idea that nodes that are close in the embedding space should have a high probability of being connected.

The encoder is a standard GNN with multi-head dot-product attention, that uses the adjacency matrix $A$ as the edge structure (and no additional edge features). In order to break symmetry, we need some way to distinguish the nodes from each other. If we are just interested in learning structure, then we do not have access to node features, only the adjacency matrix. In this case, we generate node features $H$ using random Gaussian variables $\bh_i \sim \neighborhood(0, \sigma^2I)$, where we use $\sigma^2=0.3$. This allows the graph network to learn how to appropriately separate and cluster nodes according to $A$. We generate a new set of random features each time we encode a graph. This way, the graph can only rely on the features to break symmetry, and must rely on the graph structure to generate a useful encoding.

Putting the GNF together with the graph encoder, we map training graphs from $H$ to $X$ and use this as training inputs for the GNF. Generating involves sampling $Z\sim \neighborhood(0, I)$ followed by inverting the GNF, $X=f^{-1}(Z)$, and finally decoding $X$ into $A$ and thresholding to get binary edges.
\vspace{-0.1in}

%% file: sections/experiments.tex
\section{Supervised Experiments}
\vspace{-0.1in}
In this section we study the capabilities of the supervised GNF, the GrevNet architecture.

\textbf{Datasets/Tasks: } We experiment on two types of tasks. Transductive learning tasks consist of semi-supervised document classification in citation networks (Cora and Pubmed datasets), where we test our model with the author's dataset splits~\cite{yang16_citation_data}, as well as 1\% train split for a fair comparison against~\cite{liao18_rbp}. Inductive Learning tasks consist of PPI (Protein-Protein Ineraction Dataset)~\cite{Zitnik2017_ppidata} and QM9 Molecule property prediction dataset~\cite{rama14_qm9}. For transductive learning tasks we report classification accuracy, for PPI we report Micro F1 score, and for QM9, Mean Absolute Error (MAE). More details on datasets are provided in the supplementary material.

\textbf{Baselines:} We compare GRevNets to: (1) A vanilla GNN architecture with an identical architecture and the same number of message-passing steps; (2) Neumann-RBP~\cite{liao18_rbp} -- which, to the best of our knowledge, is the state-of-the-art in the domain of memory-efficient GNNs.

\subsection{Performance on benchmark tasks}

Table~\ref{tab:grevnet-all} compares GRevNets to GNNs and Neumann RBP (NRBP). 1\% train uses 1\% of the data for training to replicate the settings in~\cite{liao18_rbp}. For these, we provide average numbers for GNN and GRevNet and \textbf{best} numbers for NRBP. The GRevNet architecture is competitive with a standard GNN, and outperforms NRBP.

\begin{table}[ht]
\begin{center}
     \begin{tabular}{clrrr}
     \toprule
     \multicolumn{2}{c}{\textbf{Dataset/Task}} & \textbf{GNN} & \textbf{GRevNet} & \textbf{Neumann RBP} \\
     \midrule
     Cora & Semi-Supervised & 71.9 & \textbf{74.5}  &  56.5 \\
     Cora & (1\% Train) & 55.5 & \textbf{55.8} & 54.6 \\
     \midrule
     Pubmed & Semi-Supervised & \textbf{76.3} & 76.0 & 62.4 \\
     Pubmed & (1\% Train) & 76.6 & \textbf{77.0} & 58.5 \\
     \midrule
     PPI & Inductive & \textbf{0.78} & 0.76 & 0.70 \\
     \bottomrule
    \end{tabular}
\end{center}
\vspace{-0.2cm}

\begin{center}
    \begin{tabular}{crrrrrrrrrrrr}

     \toprule
     \textbf{Model} & \textbf{mu} & \textbf{alpha} & \textbf{HOMO} & \textbf{LUMO} & \textbf{gap} & \textbf{R2} \\
     \midrule
     GNN & 0.474 & 0.421 & \textbf{0.097} & 0.124 & 0.170 & 27.150 \\
     GrevNet & \textbf{0.462} & \textbf{0.414} & 0.098 & 0.124 & \textbf{0.169} & \textbf{26.380} \\
     \midrule
     \midrule
     \textbf{Model} & \textbf{ZPVE} & \textbf{U0} & \textbf{U} & \textbf{H} & \textbf{G} & \textbf{Cv} \\
     \midrule
     GNN & 0.035 & 0.410 & \textbf{0.396} & \textbf{0.381} & 0.373 & 0.198 \\
     GrevNet & 0.036 & \textbf{0.390} & 0.407 & 0.418 & \textbf{0.359} & \textbf{0.195} \\
     \bottomrule
    \end{tabular}
\end{center}
    \caption{{Top: performance in terms of accuracy (Cora, Pubmed) and Micro F1 scores (PPI). For GNN and GrevNet, number of MP steps is fixed to 4. For Neumann RBP, we use 100 steps of MP. These values are averaged out over 3-5 runs with different seeds. Bottom: performance in terms of Mean Absolute Error (lower is better) for independent regression tasks on QM9 dataset. Number of MP steps is fixed to 4. The model was trained for 350k steps, as in~\cite{gilmer17_mpnn}.}}
\label{tab:grevnet-all}
\end{table}

\section{Unsupervised Experiments}
\subsection{Structured Density Estimation}
\begin{figure*}[!ht]
\begin{subfigure}{.33\textwidth}
   \includegraphics[width=\textwidth]{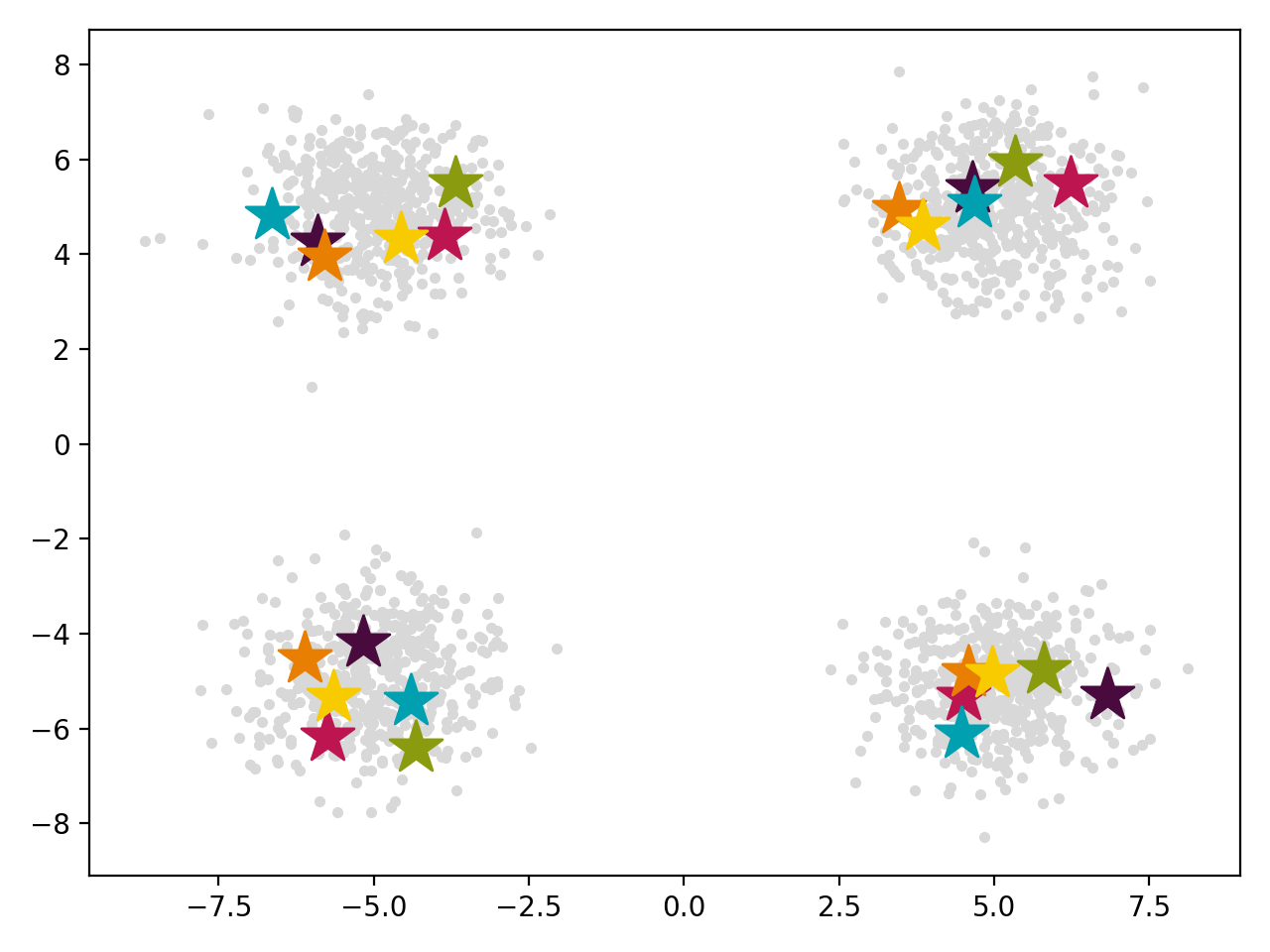}
   \caption{Training examples}
\label{fig:grevnet_training_data}
\end{subfigure}
\begin{subfigure}{.33\textwidth}
   \includegraphics[width=\textwidth]{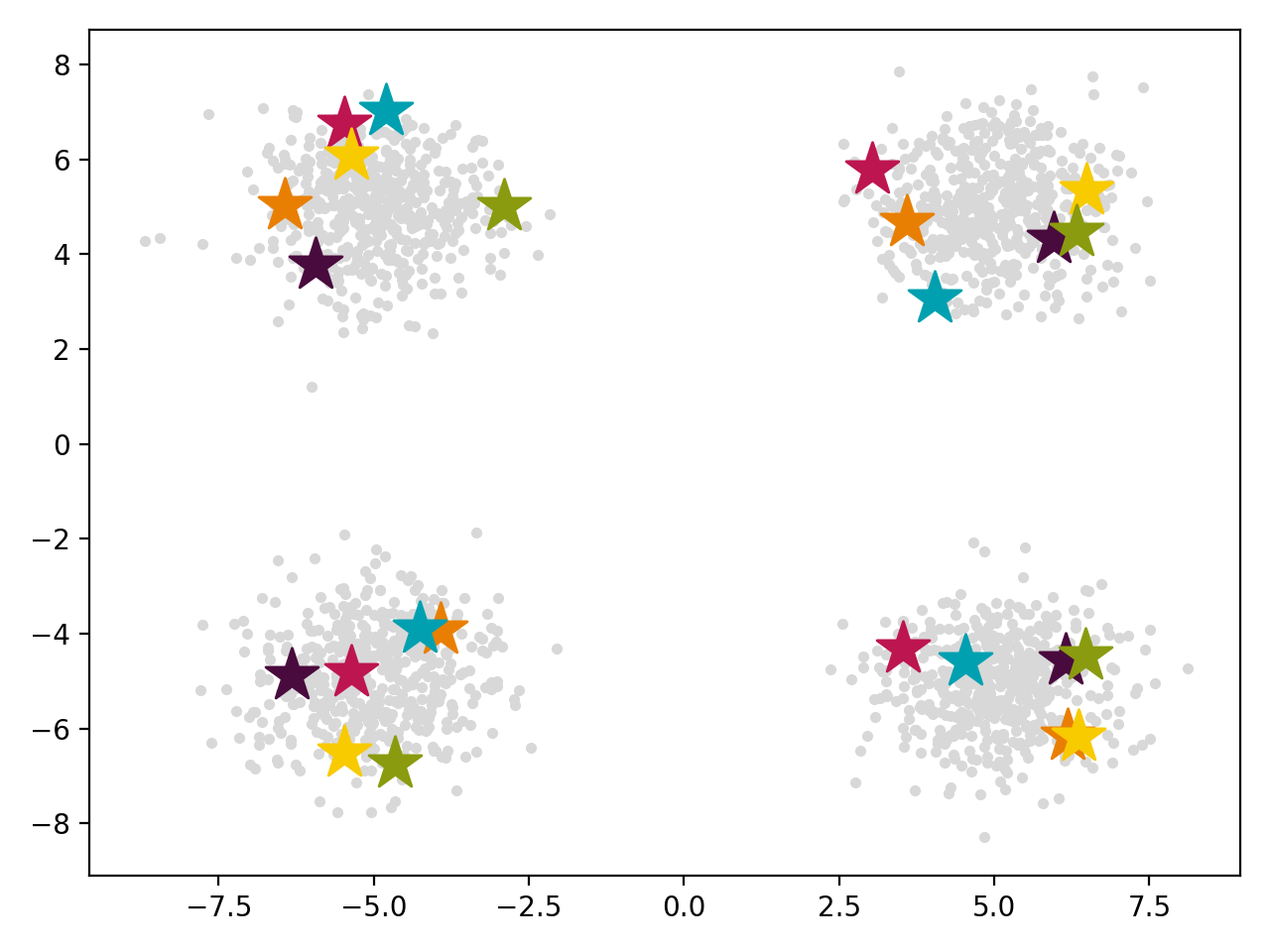}
   \caption{GNF samples}
\end{subfigure}
\begin{subfigure}{.33\textwidth}
   \includegraphics[width=\textwidth]{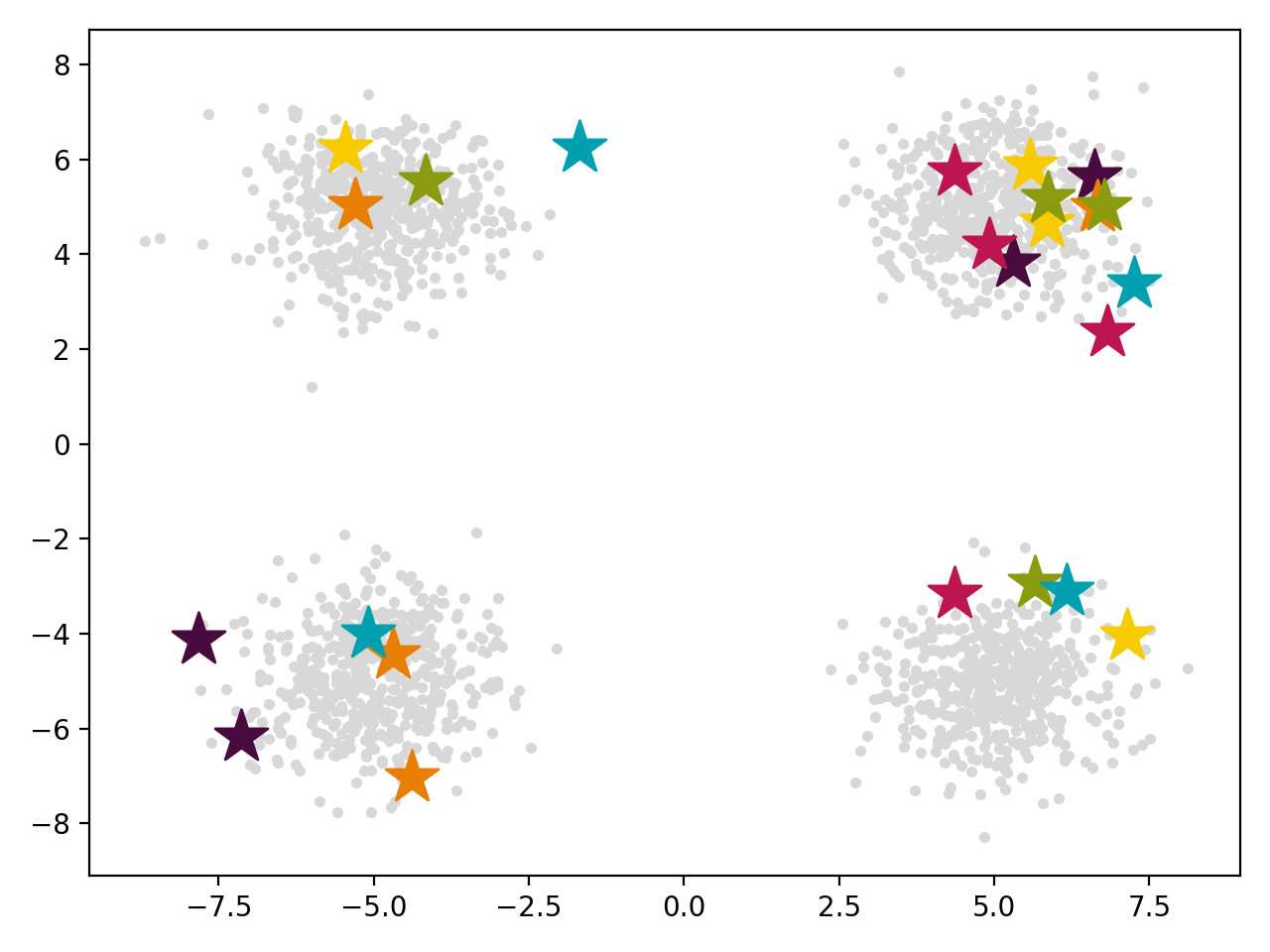}
   \caption{RealNVP samples}
\end{subfigure}
\caption{(a) shows the aggregate training distribution for the \textsc{MoG} dataset in gray, as well as 5 individual training examples. Each training example is shown in a different color and is a structured set of nodes where each node is drawn from a different Gaussian. (b) and (c) each show 5 generated samples from GNF and RealNVP, selected randomly. Each sample is shown in a different color. Note that, GNF learns to generate structured samples where each node resembles a sample from a different Gaussian, while RealNVP by design cannot model these dependencies. Best viewed in color.}
\label{fig:mog-samples}
\vspace{-0.0cm}
\end{figure*}

We compare the performance of GNFs with RealNVP for structured density estimation on 3 datasets. Details of the model architecture can be found in the supplementary material.
\paragraph{Datasets.}The first dataset is \textsc{Mixture of Gaussians (MoG)}, where each training example is a set of 4 points in a square configuration. Each point is drawn from a separate isotropic Gaussian, so no two points should land in the same area. \textsc{Mixture of Gaussians Ring (MoG Ring)} takes each example from \textsc{MoG} and rotates it randomly about the origin, creating an aggregate training distribution that forms a ring. \textsc{6-Half Moons} interpolates the original half moons dataset using 6 points with added noise.
\vspace{-0.2cm}
\paragraph{Results.} Our results are shown in Table \ref{tab:grevnet_synthetic}. We outperform \realnvp{} on all three datasets. We also compare the generated samples of the two models on the \textsc{MoG} dataset in Figure~\ref{fig:mog-samples}.

\begin{table*}
\vspace{-0.5cm}
\label{grevnet_results}
\vskip 0.15in
\begin{center}
\begin{small}
\begin{sc}
\begin{tabular}{lrrrr}
\toprule
Model & MoG (NLL) & MoG Ring (NLL) & 6-Half Moons (NLL) \\
\midrule
RealNVP & 4.2 & 5.2 & -1.2   \\
GNF & \textbf{3.6} & \textbf{4.2} & \textbf{-1.7} \\
\bottomrule
\end{tabular}
\end{sc}
\end{small}
\end{center}
\vskip -0.1in
\caption{Per-node negative log likelihoods (NLL) on synthetic datasets for \realnvp{} and \gnf{}.}
\label{tab:grevnet_synthetic}
\vspace{-0.5cm}
\end{table*}

\subsection{Graph Generation}
\paragraph{Baselines.} We compare our graph generation model on two datasets, \textsc{community-small} and \textsc{ego-small} from GraphRNN~\cite{you18graphrnn}. \textsc{community-small} is a procedurally-generated set of 100 2-community graphs, where $12 \leq |V| \leq 20$. \textsc{Ego-small} is a set of 200 graphs, where $4 \leq |V| \leq 18$, drawn from the larger Citeseer network dataset~\cite{Sen08collectiveclassification}.  For all experiments described in this section, we used scripts from the GraphRNN codebase \cite{graphrnncode} to generate and split the data. 80\% of the data was used for training and the remainder for testing.

\subsubsection{Graph Auto-Encoder}
We first train a graph auto-encoder with attention, as described in Section~\ref{sec:graph-auto-encoders}. Every training epoch, we generate new Gaussian noise features for each graph as input to the encoder.
The GNN consists of 10 MP steps, where each MP step uses a self-attention module followed by a multi-layer perceptron. Additional details can be found in the supplementary material.

Table \ref{tab:encoder_decoder_results} shows that our auto-encoder generalizes well to unseen test graphs, with a small gap between train and test cross-entropy. The total \# of incorrect edges metric shows that the model achieves good test reconstruction on \textsc{ego-small} and near-perfect test reconstruction on \textsc{community-small}.

\begin{table*}
\begin{center}
\begin{sc}
\begin{tabular}{lcccccc}
\cmidrule{2-7}
&\multicolumn{2}{c}{Binary CE}&\multicolumn{2}{c}{Total \# Incorrect Edges}&\multicolumn{2}{c}{Total \# Edges} \\
\midrule
Dataset & Train & Test & Train & Test & Train & Test  \\
\midrule
\midrule
ego-small & 9.8e-4  & 11e-04 & 24  & 32 & 3758 & 984 \\
community-small & 5e-4 & 7e-04 & 10  & 2 & 1329 & 353 \\
\bottomrule
\end{tabular}
\end{sc}
\end{center}
\vskip -0.1in
\caption{Train and test binary cross-entropy (CE) as described in equation \ref{eqn:graph_autoencoder_loss}, averaged over the total number of nodes. \textsc{Total \# Incorrect Edges} measures the number of incorrect edge predictions (either missing or extraneous) in the reconstructed graphs over the entire dataset. \textsc{Total \# Edges} lists the total number of edges in each dataset. As we use Gaussian noise for initial node features, we averaged 5 runs of our model to obtain these metrics.}
\label{tab:encoder_decoder_results}
\vspace{-0.2cm}
\end{table*}

\subsubsection{Graph Normalizing Flows for Permutation Invariant Graph Generation}
Our trained auto-encoder gives us a distribution over node embeddings that are useful for graph reconstruction. We then train a GNF to maximize the likelihood of these embeddings using an isotropic Gaussian as the prior. Once trained, at generation time the model flows $N$ random Gaussian embeddings sampled from the prior to $N$ node embeddings that describe a graph adjacency when run through the decoder.

Our GNF consists of 10 MP steps with attention and an MLP for each of $F_1$, $F_2$, $G_1$, and $G_2$.
For more details on the architecture see the supplementary material.

\paragraph{Evaluating Generated Graphs.}
We evaluate our model by providing visual samples and by using the quantitative evaluation technique in GraphRNN~\cite{you18graphrnn}, which calculates the MMD  distance ~\cite{gretton2012kernel} between the generated graphs and a previously unseen test set on three statistics based on degrees, clustering coefficients, and orbit counts. We use the implementation of GraphRNN provided by the authors to train their model and their provided evaluation script to generate all quantitative results.

In~\cite{graphrnncode}, the MMD evaluation was performed by using a test set of $N$ ground truth graphs, computing their distribution over $|V|$, and then searching for a set of $N$ generated graphs from a much larger set of samples from the model that closely matches this distribution over $|V|$. These results tend to exhibit considerable variance as the graph test sets were quite small.

To achieve more certain trends, we also performed an evaluation by generating 1024 graphs for each model and computing the MMD distance between this generated set of graphs and the ground truth test set. We report both evaluation settings in Table \ref{tab:graph_gen}. We also report results directly from \cite{you18graphrnn} on two other graph generation models, \graphvae{} and \deepgmg{}, evaluated on the same graph datasets.

\begin{figure}[h!]
\begin{subfigure}{.32\textwidth}
   \includegraphics[trim={0 35.12cm 0 0},clip,width=\textwidth]{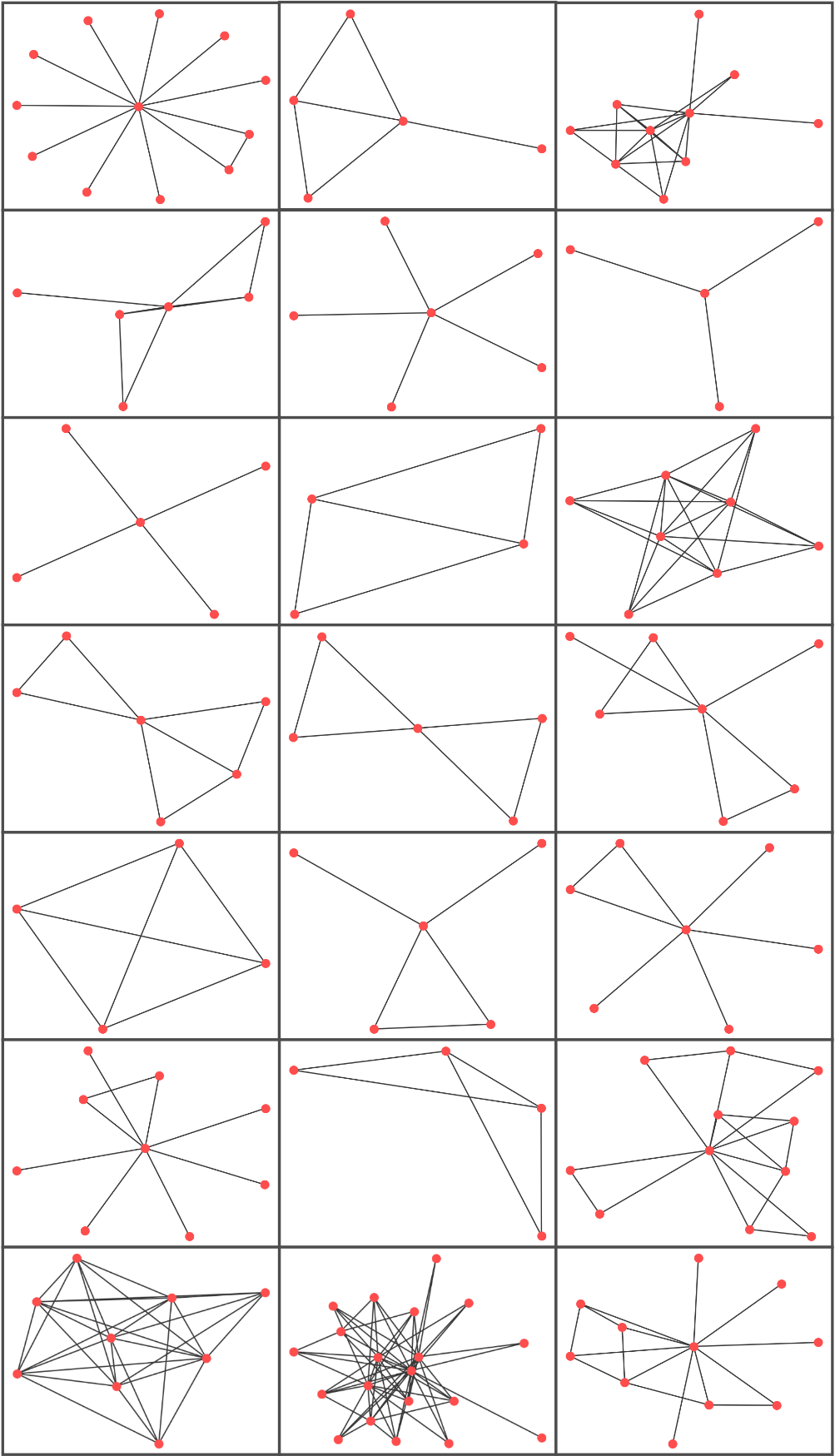}
\end{subfigure}
\hfill
\begin{subfigure}{.32\textwidth}
   \includegraphics[trim={0 35.12cm 0 0},clip,width=\textwidth]{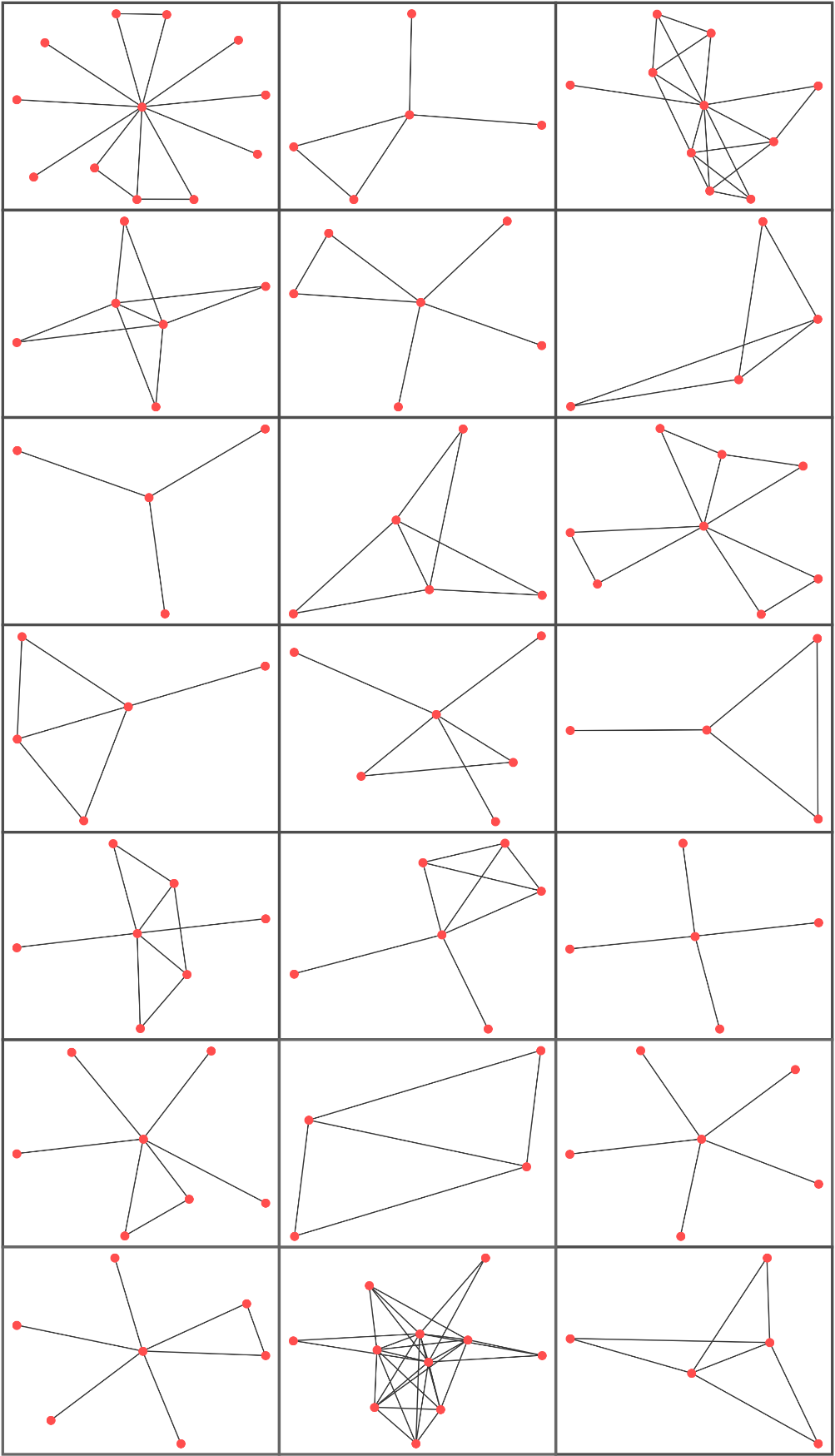}
\end{subfigure}
\hfill
\begin{subfigure}{.32\textwidth}
   \includegraphics[trim={0 35.12cm 0 0},clip,width=\textwidth]{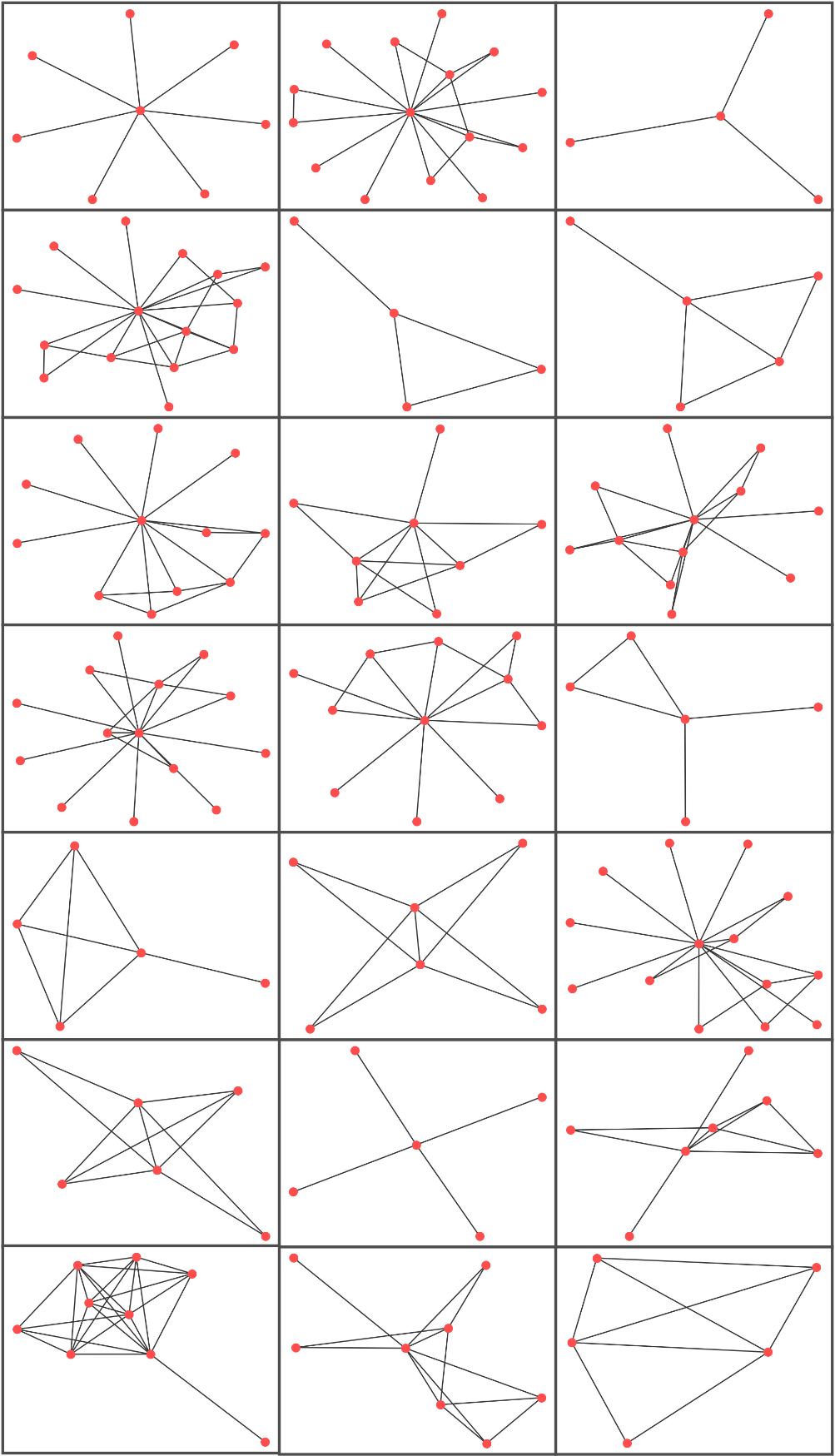}
\end{subfigure} \\

\vspace{0.25cm}
\begin{subfigure}{.32\textwidth}
   \includegraphics[trim={0 35.12cm 0 0},clip,width=\textwidth]{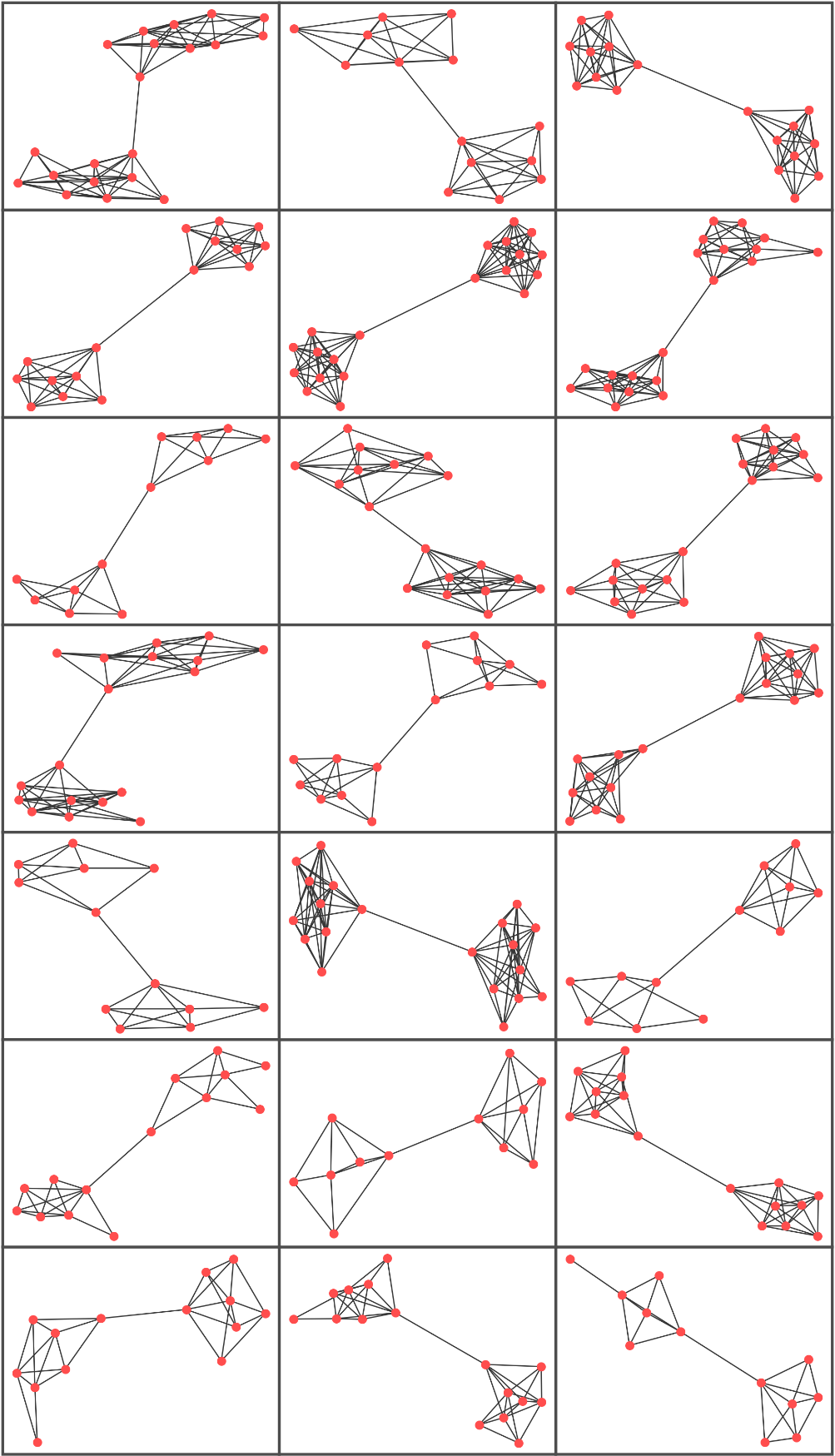}
   \caption{Training data}
\end{subfigure}
\hfill
\begin{subfigure}{.32\textwidth}
   \includegraphics[trim={0 35.12cm 0 0},clip,width=\textwidth]{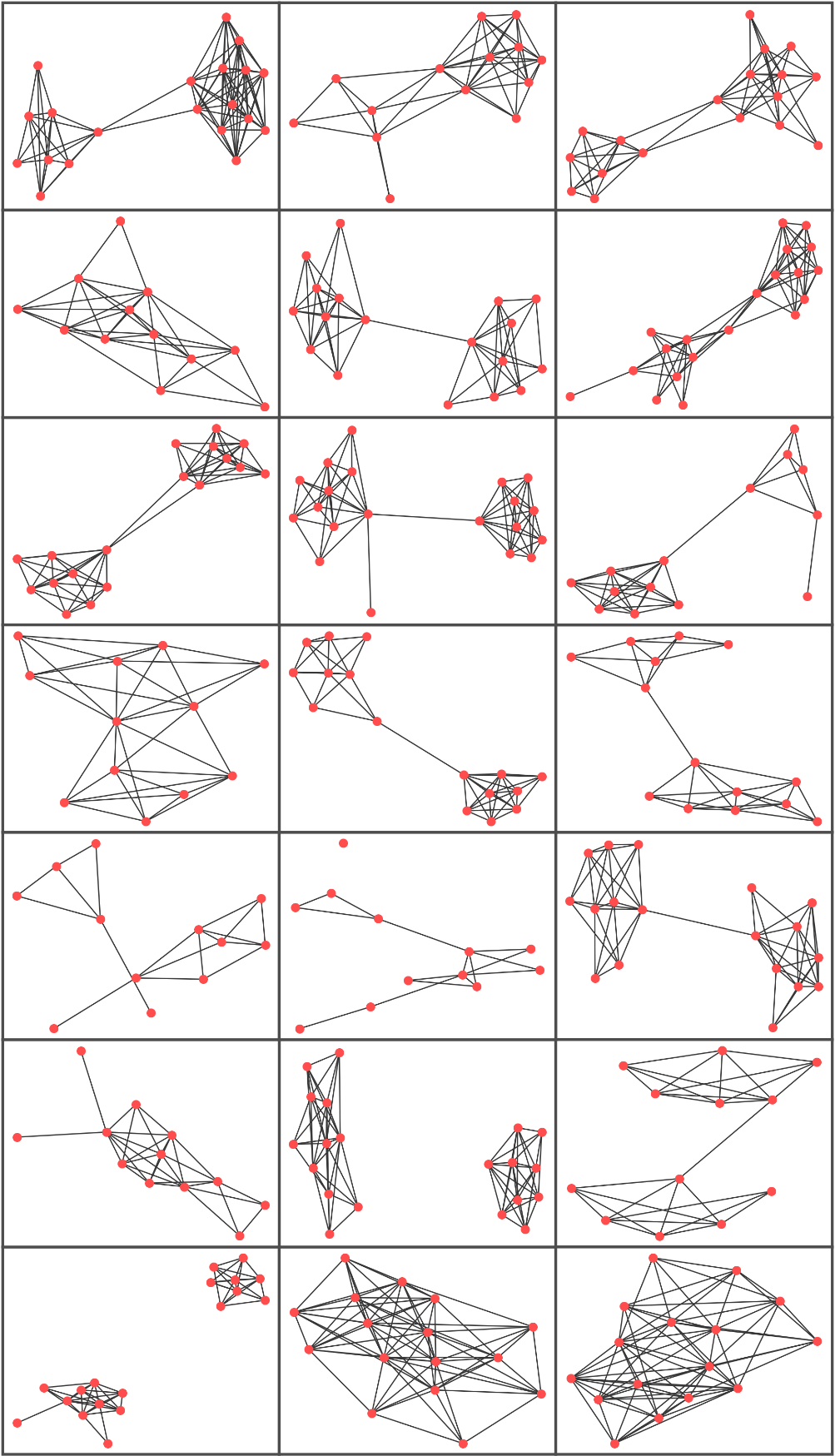}
   \caption{\textsc{GNF} samples}
\end{subfigure}
\hfill
\begin{subfigure}{.32\textwidth}
   \includegraphics[trim={0 35.12cm 0 0},clip,width=\textwidth]{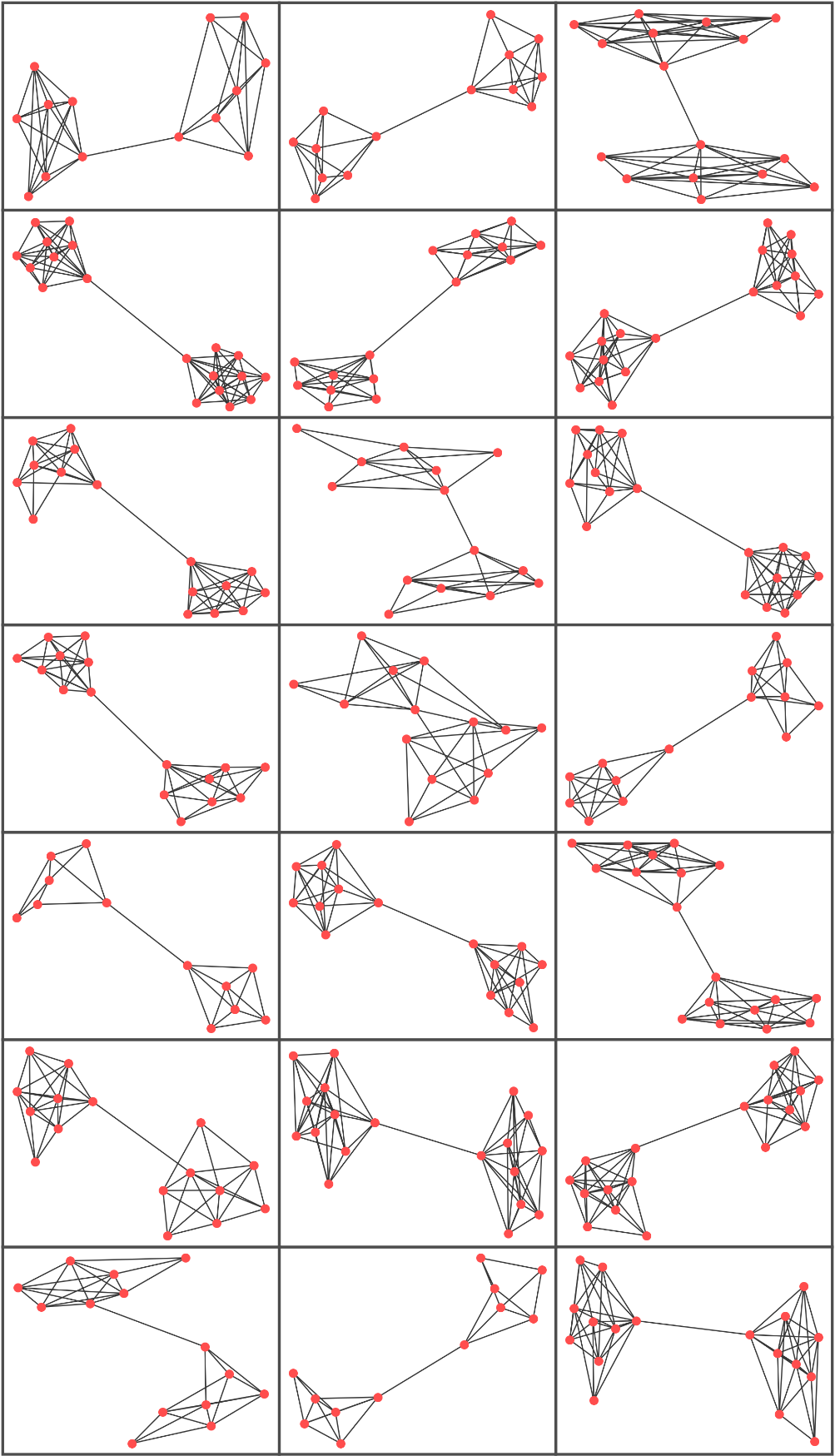}
   \caption{\textsc{GraphRNN} samples}
\end{subfigure}
\caption{Dataset examples and samples, drawn randomly, from the generative models. Top row: \ego{}, bottom row: \community{}.}
\label{fig:graph_gen}
\vspace{-0.5cm}
\end{figure}

\paragraph{Results.}
\vspace{-0.1in}
We provide a visualization of generated graphs from \graphrnn{} and \gnf{} in Figure~\ref{fig:graph_gen}. As shown in Table \ref{tab:graph_gen}, \gnf{} outperforms \graphvae{} and \deepgmg{}, and is competitive with \graphrnn{}. Error margins for \gnf{} and \graphrnn{} and a larger set of visualizations are provided in the supplementary material.
 \begin{table*}
\begin{center}
\begin{sc}
\begin{tabular}{lcccccc}
\cmidrule{2-7}
&\multicolumn{3}{c}{\textbf{\community}}&\multicolumn{3}{c}{\textbf{\ego}}\\
\midrule
\textbf{Model} & Degree& Cluster & Orbit & Degree & Cluster & Orbit  \\
\midrule
\graphvae & 0.35 & 0.98 & 0.54 & 0.13 & 0.17 & 0.05 \\
\deepgmg & 0.22 & 0.95 & 0.4  & 0.04 & 0.10 & 0.02 \\
\midrule
\midrule
\graphrnn &0.08 & 0.12 &0.04 & 0.09 &0.22 &0.003 \\
\gnf &0.20 & 0.20 &0.11 & 0.03 &0.10 &0.001 \\
\midrule
\midrule
\graphrnn (1024) &\textbf{0.03 }& \textbf{0.01 }&\textbf{0.01 }& 0.04 &0.05 &0.06 \\
\gnf (1024) &0.12 &0.15 &0.02 & \textbf{0.01 }&\textbf{0.03 }&\textbf{0.0008} \\
\bottomrule
\end{tabular}
\end{sc}

\end{center}
\vskip -0.1in
\caption{ Graph generation results depicting MMD for various graph statistics between the test set and generated graphs. \graphvae{} and \deepgmg{} are reported directly from~\cite{you18graphrnn}. The second set of results (\graphrnn{}, \gnf{}) are from evaluating the GraphRNN evaluation scheme with node distribution matching turned on. We trained 5 separate models of each type and performed 3 trials per model, then averaged the result over 15 runs. The third set of results (\graphrnn{} (1024), \gnf{} (1024)) are obtained when evaluating on the test set over all 1024 generated graphs (no sub-sampling of the generated graphs based on node similarity). In this case, we trained and evaluated the result over 5 separate runs per model.}
\label{tab:graph_gen}
\end{table*}

%% file: sections/conclusion.tex
\section{Conclusion}
\vspace{-0.2cm}
We propose GNFs, normalizing flows using GNNs based on the RealNVP, by making the message passing steps reversible. In the supervised case, reversibility allows for backpropagation without the need to store hidden node states. This provides significant memory savings, further pushing the scalability limits of GNN architectures. On several benchmark tasks, GNFs match the performance of GNNs, and outperform Neumann RBP. In the unsupervised case, GNFs provide a flexible distribution over a set of continuous vectors. Using the pre-trained embeddings of a novel graph auto-encoder, we use GNFs to learn a distribution over the embedding space, and then use the decoder to generate graphs. This model is permutation invariant, yet competitive with the state-of-the-art auto-regressive GraphRNN model. Future work will focus on applying GNFs to larger graphs, and training the GNF and auto-encoder in an end-to-end approach.

%% file: sections/supplementary_material.tex
\section{Supplementary Material}
\label{sec:supplementary}
\subsection{Supervised Experiments Details}
\subsubsection{Datasets Description}
\label{sec:dataset}
The following datasets were used in the experiments we reported in the main body of the paper.
\begin{itemize}
     \item \emph{Molecule property prediction on QM9}~\cite{rama14_qm9}: 
    which consists of about 134k drug-like molecules made up of Hydrogen (H), Carbon (C), Oxygen (O), Nitrogen (N), and Flourine (F) atoms containing up to 9 heavy (non Hydrogen) atoms.
    \item \emph{Semi-supervised document classification on citation networks}: A node of a network represents a document associated with a bag-of-words feature. Nodes are connected based on the citation links. Given a portion of nodes labeled with subject categories, e.g., science, history, the task is to predict the categories for unlabeled nodes within the same network. We use two citation networks from \cite{yang16_citation_data} -  Cora and Pubmed. We try this with two settings - one with the author provided  dataset splits into train/test/validation and the other with 1\%/49\%/50\% train/test/validation splits.
    \item \emph{Inductive Learning on Protein-Protein Interaction (PPI) Dataset}: PPI consists of graphs corresponding to different human tissues~\cite{Zitnik2017_ppidata}. The dataset contains 20 graphs for training, 2 for validation and 2 for testing. Testing graphs remain completely unobserved during training. To construct the graphs, we used the preprocessed data provided by \cite{hamilton17_inductive} and \cite{v2018_gat}.
\end{itemize}

\subsubsection{Hyperparameter Tuning and Other Details}
\label{sec:hparams}
The following list describes major hyperparameter settings and some other implementation details for our model.
\begin{itemize}
    \item \emph{L2 parameter regularization:} In all models, for all runs, we applied a L2-regularization on the weights with a coefficient of 0.001. 
    \item \emph{Number of message passing steps:} We performed a search for the number of message passing steps over the following set - [1, 2, 4, 5, 10, 20] on Cora and Pubmed. We found that 4 works the best for GNN and GRevNet, and stuck to that for experiments on all datasets. For Neumann RBP, we tried 100 and 200 message passing steps, of which 100 worked better.
    \item \emph{Selection of the test model:} We selected the test model by storing the model with the best performance in terms of accuracy/Micro F1 score/Mean \emph{Squared} error (for QM9) on a held-out validation dataset.
    \item \emph{Batch Normalization:} It is observed that as the number of message passing steps increases beyond a limit (in our case it was 20), the GNN/GRevNet model starts to perform worse, and often it is hard to optimize the whole system well -- in an end-to-end manner. Likely, the whole model ends up at a bad optimum and is unable to recover from it. Similar observations were made by ~\cite{Kipf2016_gcn}. In order to tackle this problem, we applied batch norm at each of the layers during message passing. This helps with training up to about 40 steps. In the results with, 4 and 10 steps of message passing, we don't use Batch Normalization.
    \item \emph{Optimization:} We used Adam Optimizer~\cite{kingma14_adam} for optimizing GNNs and GRevNets. We chose a fixed learning rate of 1e-4. Changing the learning rate to 1e-3, sometimes doesn't work and training is unstable. We applied gradient clipping, allowing a maximum gradient norm of 4.0 in all cases. For QM9, we chose a learning rate of 1e-3, as the authors specify in the MPNN paper~\cite{gilmer17_mpnn}. For Neumann RBP, we found that Adam doesn't work well. So, we chose the settings specified by the author, that is SGD with Momentum of 0.9 and a learning rate of 1e-3.    
    \item \emph{Architecture Design: } For the message generation step of the message passing phase, we use an MLP over the node features. For the update step during message passing, we use a GRU-like update to update the node features. The final classifier/regressor on top of the graph net module was an MLP with 2 layers.
\end{itemize}


\subsection{Unsupervised Experiment Details}
\subsubsection{Results with Error Bars}
In Table~\ref{tab:graph_gen_supp}, we show the results with error bars for GraphRNN and GNF. GraphVAE and DeepGMG are reported directly from~\cite{you18graphrnn}.
\begin{table*}
\scriptsize
\begin{center}
\begin{sc}
\begin{tabular}{lcccccc}
\cmidrule{2-7}
&\multicolumn{3}{c}{\textbf{\community}}&\multicolumn{3}{c}{\textbf{\ego}}\\
\midrule
\textbf{Model} & Degree& Cluster & Orbit & Degree & Cluster & Orbit  \\
\midrule
\graphvae & 0.35 & 0.98 & 0.54 & 0.13 & 0.17 & 0.05 \\
\deepgmg & 0.22 & 0.95 & 0.4  & 0.04 & 0.10 & 0.02 \\
\midrule
\midrule
\graphrnn &0.08 $\pm$ 0.06& 0.12$\pm$ 0.07&0.04 $\pm$0.04 & 0.09$\pm$ 0.10&0.22$\pm$0.16 &0.003$\pm$0.004 \\
\gnf &0.20 $\pm$ 0.07&0.20$\pm$0.07 &0.11$\pm$ 0.07& 0.03 $\pm$ 0.03 &0.10 $\pm$ 0.05&0.001 $\pm$ 0.0009 \\
\midrule
\midrule
\graphrnn (1024) &\textbf{0.03 $\pm$ 0.02}& \textbf{0.01$\pm$ 0.0007}&\textbf{0.01 $\pm$0.009 }& 0.04$\pm$ 0.02&0.05$\pm$0.02 &0.06$\pm$0.05 \\
\gnf (1024) &0.12 $\pm$ 0.006&0.15$\pm$0.004 &0.02$\pm$ 0.003& \textbf{0.01 $\pm$ 0.003 }&\textbf{0.03 $\pm$ 0.004}&\textbf{0.0008 $\pm$ 0.0002} \\
\bottomrule 
\end{tabular}
\end{sc}

\end{center}
\vskip -0.1in
\caption{Graph generation results showing MMD for various graph statistics between the test set and generated graphs. \graphvae{} and \deepgmg{} are reproduced directly from the GraphRNN paper. The second set of results (\graphrnn{}, \gnf{}) are from running the GraphRNN evaluation script with node distribution matching turned on. We trained 5 separate models of each type and did 3 runs per models, then took the average over the 15 runs. The third set of results (\graphrnn{} (1024), \gnf{} (1024)) are from evaluating on the test set and 1024 generated graphs. Again we trained 5 separate models of each type and evaluated the MMD over 5 separate runs, 1 run per model.}
\label{tab:graph_gen_supp}
\end{table*}
\subsubsection{More Graph Samples}
\begin{figure}[h!]
\begin{subfigure}{.32\textwidth}
   \includegraphics[width=\textwidth]{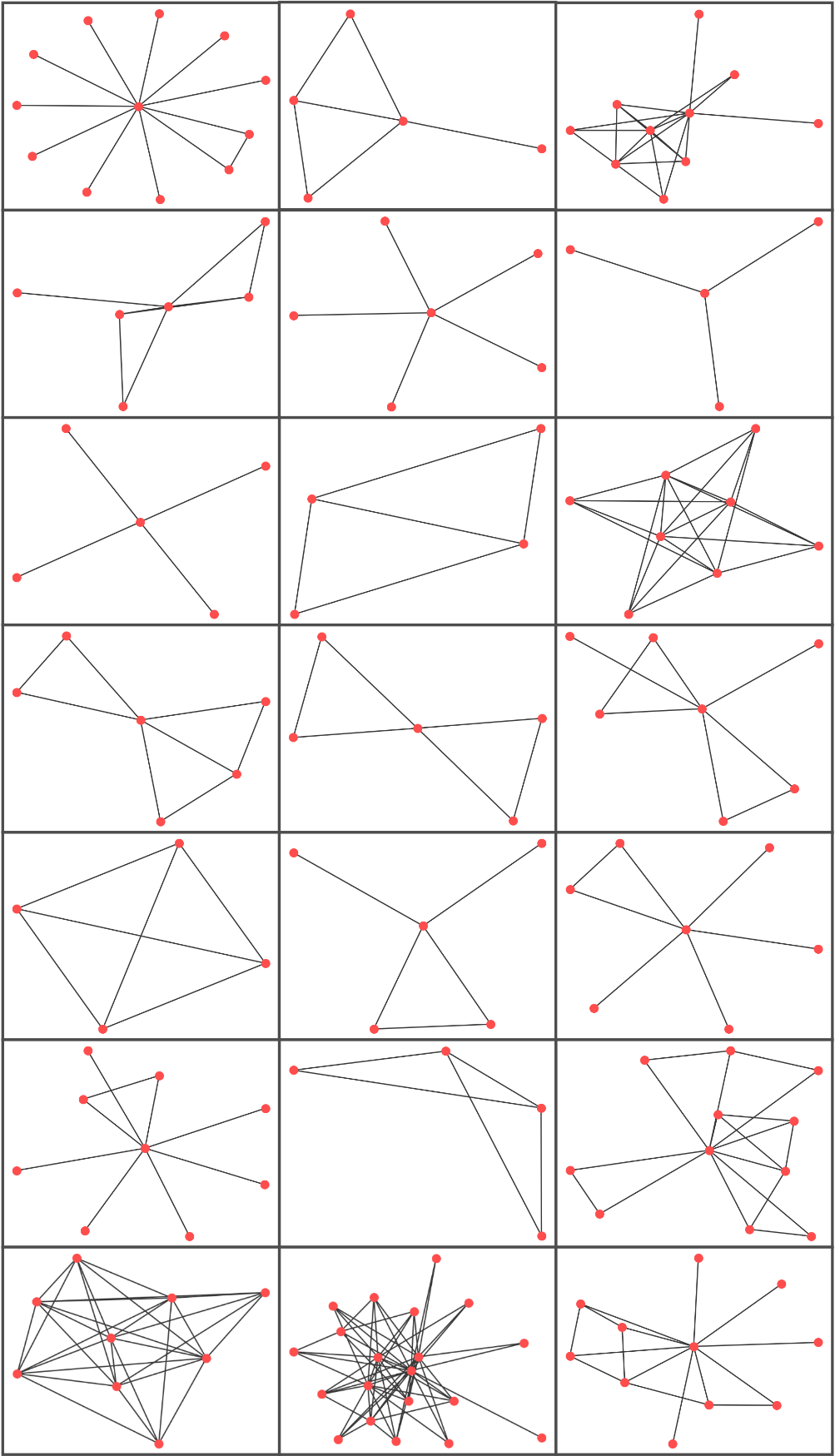}
   \caption{Training data}
\end{subfigure}
\hfill
\begin{subfigure}{.32\textwidth}
   \includegraphics[width=\textwidth]{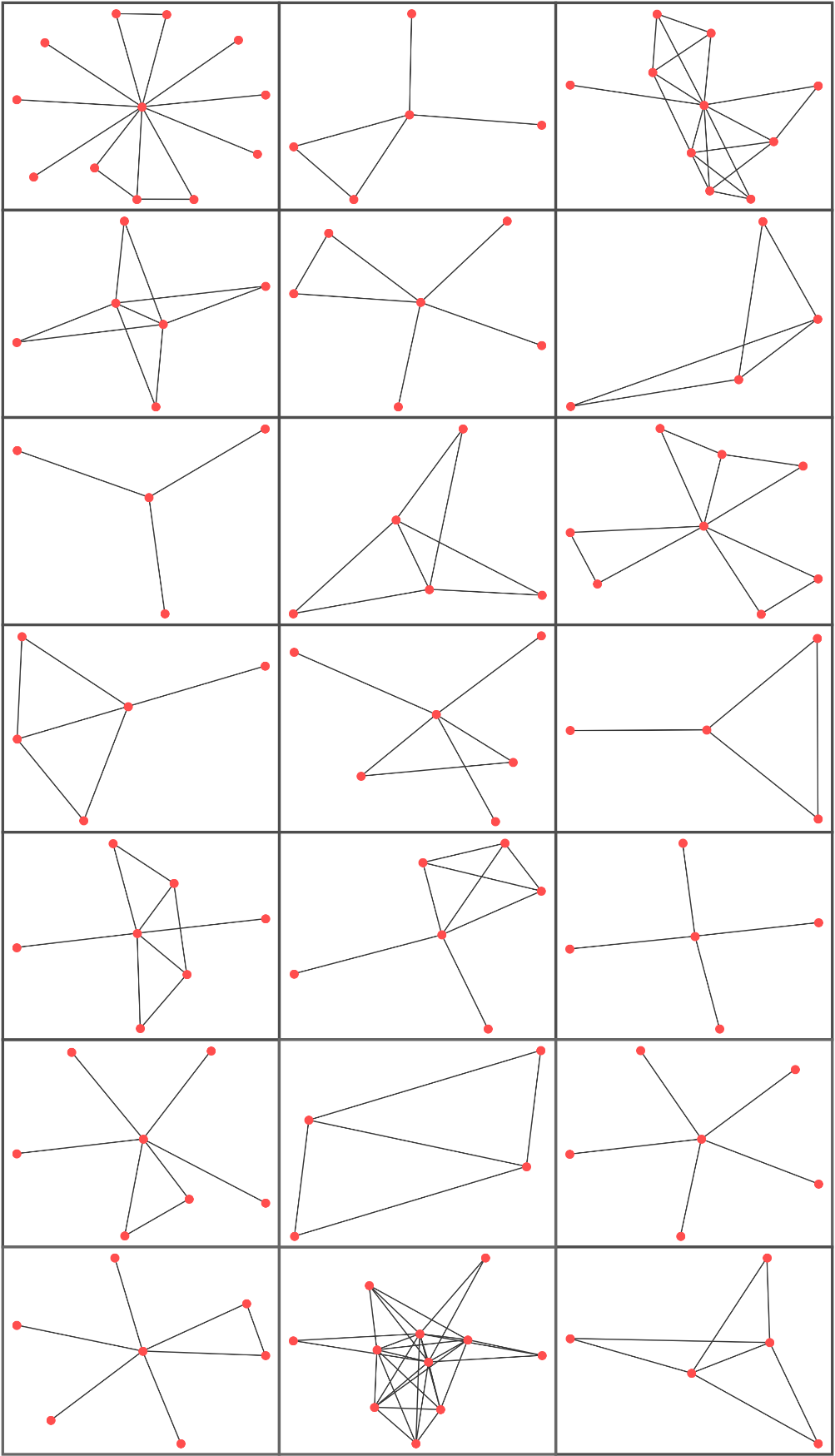}
   \caption{\textsc{GNF} samples}
\end{subfigure}
\hfill
\begin{subfigure}{.32\textwidth}
   \includegraphics[width=\textwidth]{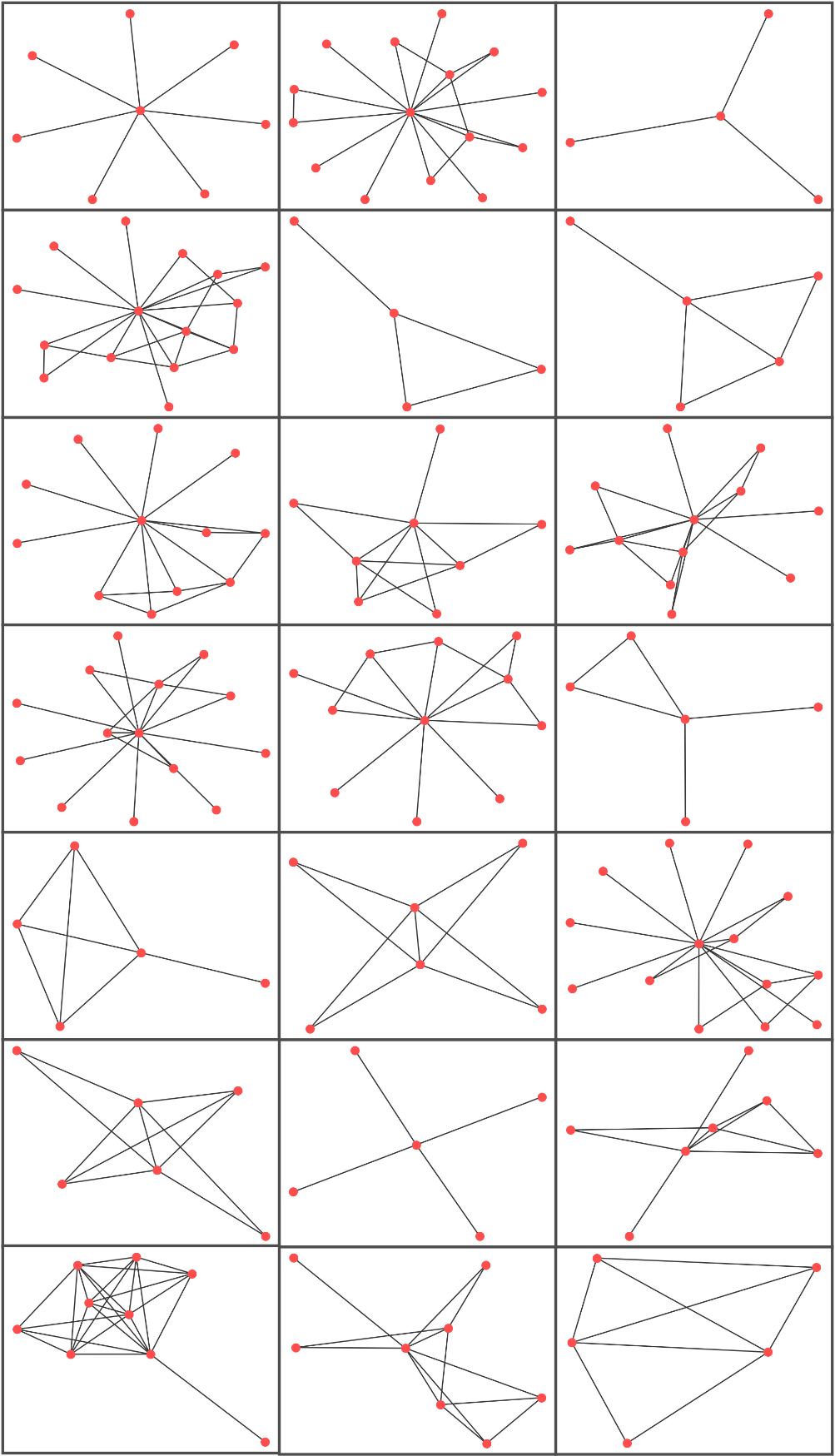}
   \caption{\textsc{GraphRNN} samples}
\end{subfigure}
\caption{Left, training data graphs from \ego{}. Middle, generated graphs from \gnf{}. Right, generated graphs from \graphrnn{}. Samples were picked at random.}
\label{fig:ego_graph_gen}
\end{figure}

\begin{figure}[h!]
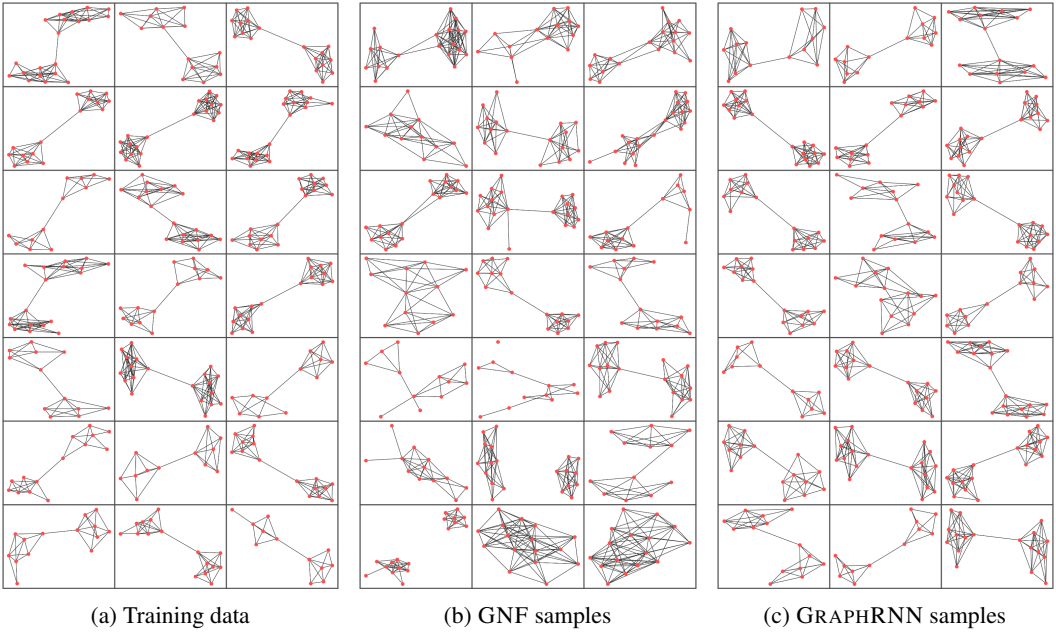

\begin{subfigure}{.32\textwidth}
   \includegraphics[width=\textwidth]{graph_imgs/community_truth.png}
   \caption{Training data}
\end{subfigure}
\hfill
\begin{subfigure}{.32\textwidth}
   \includegraphics[width=\textwidth]{graph_imgs/community_grevnet.png}
   \caption{\textsc{GNF} samples}
\end{subfigure}
\hfill
\begin{subfigure}{.32\textwidth}
   \includegraphics[width=\textwidth]{graph_imgs/community_graph_rnn_good.png}
   \caption{\textsc{GraphRNN} samples}
\end{subfigure}
\caption{Left, training data graphs from \community{}. Middle, generated graphs from \gnf{}. Right, generated graphs from \graphrnn{}. Samples were picked at random.}
\label{fig:comm_graph_gen}
\end{figure}

In Figures~\ref{fig:ego_graph_gen} and~\ref{fig:comm_graph_gen} we show the full set of samples on the \ego{} and \community{} datasets.

\subsection{Computing Infrastructure}
For all experiments in this section, we trained on a single GPU, either a Tesla P100 or Titan Xp.
\subsubsection{Structured Density Estimation}
We train a GNF with 12 message passing steps. We apply batch norm to the input at the beginning of each step, and then we use the same module for $F_1$, $F_2$, $G_1$, and $G_2$. The module consists of a dot-product multi-head self-attention layer followed by an MLP with 5 layers, latent dimension of 256, and ReLu non-linearities. We use 8 attention heads.

For RealNVP, we use an analogous architecture, with 12 coupling layers, batch norm at the beginning of each step followed by an MLP of 5 layers with latent dimension 256 and ReLu non-linearities.

We train both models for 15k steps using the Adam optimizer with a learning rate of 1e-04.

\subsubsection{Graph Auto-encoder}
We found that \ego{} and \community{} needed differing capacities for the node embedding. We used an embedding size of 14 for \ego{} and 30 for \community{}. We used 10 message passing steps. Each step uses the same architecture, a batch norm layer, followed by multi-head dot-product self-attention, and then an MLP with 3 layers, a latent dimension of 2048, and ReLu non-linearities. We used 8 attention heads. We shared weights between message passing steps. For both datasets we trained for 100k steps using the Adam Optimizer and a learning rate of 1e-04. We use an exponential learning rate decay of 0.99 every 1000 steps.

\subsubsection{GNF for Graph Generation}
 We use the same embedding sizes as the graph auto-encoder, 14 for \ego{} and 30 for \community{}. We used 12 message passing steps.  For each message passing step we used the same architecture for each $F_1$, $F_2$, $G_1$ and $G_2$. We have a batch norm layer followed by a multi-head dot-product self-attention module, then an MLP with 3 layers, a latent dimension of 2048, and ReLu non-linearities. We used 8 attention heads. We did not share weights between message passing steps. For both datasets we train for 100k steps using the Adam Optimizer and a learning rate of 1e-04 for \ego{} and 1e-05 for \community{}. We use an exponential learning rate decay of 0.99 every 1000 steps.

%% file: gnf_main.bbl
\begin{thebibliography}{31}
\providecommand{\natexlab}[1]{#1}
\providecommand{\url}[1]{\texttt{#1}}
\expandafter\ifx\csname urlstyle\endcsname\relax
  \providecommand{\doi}[1]{doi: #1}\else
  \providecommand{\doi}{doi: \begingroup \urlstyle{rm}\Url}\fi

\bibitem[Almeida(1987)]{Almeida87_rbp}
L.~B. Almeida.
\newblock A learning rule for asynchronous perceptrons with feedback in a
  combinatorial environment.
\newblock In M.~Caudil and C.~Butler, editors, \emph{Proceedings of the {IEEE}
  First International Conference on Neural Networks {San Diego, CA}}, pages
  609--618, 1987.

\bibitem[Bahdanau et~al.(2015)Bahdanau, Cho, and Bengio]{bahdanau2014neural}
Dzmitry Bahdanau, Kyunghyun Cho, and Yoshua Bengio.
\newblock Neural machine translation by jointly learning to align and
  translate.
\newblock \emph{International Conference on Learning Representations}, 2015.

\bibitem[Dinh et~al.(2014)Dinh, Krueger, and Bengio]{dinh2014nice}
Laurent Dinh, David Krueger, and Yoshua Bengio.
\newblock Nice: Non-linear independent components estimation.
\newblock \emph{arXiv preprint arXiv:1410.8516}, 2014.

\bibitem[Dinh et~al.(2017)Dinh, Sohl-Dickstein, and Bengio]{dinh2016density}
Laurent Dinh, Jascha Sohl-Dickstein, and Samy Bengio.
\newblock Density estimation using real nvp.
\newblock \emph{ICLR}, 2017.

\bibitem[Duvenaud et~al.(2015)Duvenaud, Maclaurin, Iparraguirre, Bombarell,
  Hirzel, Aspuru-Guzik, and Adams]{duvenaud2015convolutional}
David~K Duvenaud, Dougal Maclaurin, Jorge Iparraguirre, Rafael Bombarell,
  Timothy Hirzel, Al{\'a}n Aspuru-Guzik, and Ryan~P Adams.
\newblock Convolutional networks on graphs for learning molecular fingerprints.
\newblock In \emph{Advances in neural information processing systems}, pages
  2224--2232, 2015.

\bibitem[Gilmer et~al.(2017)Gilmer, Schoenholz, Riley, Vinyals, and
  Dahl]{gilmer17_mpnn}
Justin Gilmer, Samuel~S. Schoenholz, Patrick~F. Riley, Oriol Vinyals, and
  George~E. Dahl.
\newblock Neural message passing for quantum chemistry.
\newblock In \emph{International Conference on Machine Learning}, volume~70,
  pages 1263--1272, 2017.

\bibitem[Gomez et~al.(2017)Gomez, Ren, Urtasun, and Grosse]{gomez17_revnet}
Aidan~N. Gomez, Mengye Ren, Raquel Urtasun, and Roger~B. Grosse.
\newblock The reversible residual network: Backpropagation without storing
  activations.
\newblock \emph{NIPS}, 2017.
\newblock URL \url{http://arxiv.org/abs/1707.04585}.

\bibitem[Gori et~al.(2005)Gori, Monfardini, and Scarselli]{gori2005new}
Marco Gori, Gabriele Monfardini, and Franco Scarselli.
\newblock A new model for learning in graph domains.
\newblock In \emph{Proceedings. 2005 IEEE International Joint Conference on
  Neural Networks, 2005.}, volume~2, pages 729--734. IEEE, 2005.

\bibitem[Gretton et~al.(2012)Gretton, Borgwardt, Rasch, Sch\"{o}lkopf, and
  Smola]{gretton2012kernel}
Arthur Gretton, Karsten~M. Borgwardt, Malte~J. Rasch, Bernhard Sch\"{o}lkopf,
  and Alexander Smola.
\newblock A kernel two-sample test.
\newblock \emph{J. Mach. Learn. Res.}, 13:\penalty0 723--773, March 2012.
\newblock ISSN 1532-4435.
\newblock URL \url{http://dl.acm.org/citation.cfm?id=2188385.2188410}.

\bibitem[Hamilton et~al.(2017)Hamilton, Ying, and
  Leskovec]{hamilton17_inductive}
Will Hamilton, Zhitao Ying, and Jure Leskovec.
\newblock Inductive representation learning on large graphs.
\newblock In \emph{Advances in Neural Information Processing Systems}, pages
  1024--1034. 2017.

\bibitem[Kingma and Ba(2014)]{kingma14_adam}
Diederik~P. Kingma and Jimmy Ba.
\newblock Adam: A method for stochastic optimization.
\newblock \emph{CoRR}, abs/1412.6980, 2014.
\newblock URL
  \url{http://dblp.uni-trier.de/db/journals/corr/corr1412.html#KingmaB14}.

\bibitem[Kipf and Welling(2016{\natexlab{a}})]{kipf16graphvae}
Thomas~N. Kipf and Max Welling.
\newblock Variational graph auto-encoders.
\newblock \emph{CoRR}, abs/1611.07308, 2016{\natexlab{a}}.

\bibitem[Kipf and Welling(2016{\natexlab{b}})]{kipf2016semi}
Thomas~N Kipf and Max Welling.
\newblock Semi-supervised classification with graph convolutional networks.
\newblock \emph{arXiv preprint arXiv:1609.02907}, 2016{\natexlab{b}}.

\bibitem[Kipf and Welling(2017)]{Kipf2016_gcn}
Thomas~N. Kipf and Max Welling.
\newblock Semi-supervised classification with graph convolutional networks.
\newblock \emph{International Conference on Learning Representation}, 2017.

\bibitem[Li et~al.(2015{\natexlab{a}})Li, Swersky, and Zemel]{li2016gmmn}
Yujia Li, Kevin Swersky, and Rich Zemel.
\newblock Generative moment matching networks.
\newblock In \emph{Proceedings of the 32nd International Conference on Machine
  Learning}, volume~37, pages 1718--1727, 2015{\natexlab{a}}.

\bibitem[Li et~al.(2015{\natexlab{b}})Li, Tarlow, Brockschmidt, and
  Zemel]{li2015gated}
Yujia Li, Daniel Tarlow, Marc Brockschmidt, and Richard Zemel.
\newblock Gated graph sequence neural networks.
\newblock \emph{arXiv preprint arXiv:1511.05493}, 2015{\natexlab{b}}.

\bibitem[{Li} et~al.(2018){Li}, {Vinyals}, {Dyer}, {Pascanu}, and
  {Battaglia}]{li2018deepgen}
Yujia {Li}, Oriol {Vinyals}, Chris {Dyer}, Razvan {Pascanu}, and Peter
  {Battaglia}.
\newblock {Learning Deep Generative Models of Graphs}.
\newblock \emph{arXiv e-prints}, art. arXiv:1803.03324, Mar 2018.

\bibitem[Liao et~al.(2018)Liao, Xiong, Fetaya, Zhang, Yoon, Pitkow, Urtasun,
  and Zemel]{liao18_rbp}
Renjie Liao, Yuwen Xiong, Ethan Fetaya, Lisa Zhang, KiJung Yoon, Xaq Pitkow,
  Raquel Urtasun, and Richard Zemel.
\newblock Reviving and improving recurrent back-propagation.
\newblock In \emph{International Conference on Machine Learning}, volume~80,
  pages 3082--3091, 2018.

\bibitem[Makhzani et~al.(2016)Makhzani, Shlens, Jaitly, and
  Goodfellow]{makhzani2016adversarial}
Alireza Makhzani, Jonathon Shlens, Navdeep Jaitly, and Ian Goodfellow.
\newblock Adversarial autoencoders.
\newblock In \emph{International Conference on Learning Representations}, 2016.

\bibitem[Pineda(1988)]{pineda87_rbp}
Fernando~J. Pineda.
\newblock Generalization of back propagation to recurrent and higher order
  neural networks.
\newblock In D.~Z. Anderson, editor, \emph{Neural Information Processing
  Systems}, pages 602--611. American Institute of Physics, 1988.
\newblock URL
  \url{http://papers.nips.cc/paper/67-generalization-of-back-propagation-to-recurrent-and-higher-order-neural-networks.pdf}.

\bibitem[Ramakrishnan et~al.(2014)Ramakrishnan, Dral, Rupp, and von
  Lilienfeld]{rama14_qm9}
Raghunathan Ramakrishnan, Pavlo~O. Dral, Matthias Rupp, and O.~Anatole von
  Lilienfeld.
\newblock Quantum chemistry structures and properties of 134 kilo molecules.
\newblock \emph{Scientific Data}, 1:\penalty0 140022 EP --, 2014.

\bibitem[Rezende and Mohamed(2015)]{rezende15normalizing}
Danilo Rezende and Shakir Mohamed.
\newblock Variational inference with normalizing flows.
\newblock In \emph{International Conference on Machine Learning}, volume~37,
  pages 1530--1538, 2015.

\bibitem[Scarselli et~al.(2009)Scarselli, Gori, Tsoi, Hagenbuchner, and
  Monfardini]{scarselli2009graph}
Franco Scarselli, Marco Gori, Ah~Chung Tsoi, Markus Hagenbuchner, and Gabriele
  Monfardini.
\newblock The graph neural network model.
\newblock \emph{IEEE Transactions on Neural Networks}, 20\penalty0
  (1):\penalty0 61--80, 2009.

\bibitem[Sen et~al.(2008)Sen, Namata, Bilgic, Getoor, Gallagher, and
  Eliassi-Rad]{Sen08collectiveclassification}
Prithviraj Sen, Galileo Namata, Mustafa Bilgic, Lise Getoor, Brian Gallagher,
  and Tina Eliassi-Rad.
\newblock Collective classification in network data.
\newblock Technical report, 2008.

\bibitem[Vaswani et~al.(2017)Vaswani, Shazeer, Parmar, Uszkoreit, Jones, Gomez,
  Kaiser, and Polosukhin]{vaswani2017attention}
Ashish Vaswani, Noam Shazeer, Niki Parmar, Jakob Uszkoreit, Llion Jones,
  Aidan~N Gomez, {\L}ukasz Kaiser, and Illia Polosukhin.
\newblock Attention is all you need.
\newblock In \emph{Advances in Neural Information Processing Systems}, pages
  5998--6008, 2017.

\bibitem[Veličković et~al.(2018)Veličković, Cucurull, Casanova, Romero,
  Liò, and Bengio]{v2018_gat}
Petar Veličković, Guillem Cucurull, Arantxa Casanova, Adriana Romero, Pietro
  Liò, and Yoshua Bengio.
\newblock Graph attention networks.
\newblock In \emph{International Conference on Learning Representations}, 2018.
\newblock URL \url{https://openreview.net/forum?id=rJXMpikCZ}.

\bibitem[Wang et~al.(2018)Wang, Liao, Ba, and Fidler]{wang2018nervenet}
Tingwu Wang, Renjie Liao, Jimmy Ba, and Sanja Fidler.
\newblock Nervenet: Learning structured policy with graph neural networks.
\newblock \emph{ICLR}, 2018.

\bibitem[Yang et~al.(2016)Yang, Cohen, and Salakhutdinov]{yang16_citation_data}
Zhilin Yang, William~W. Cohen, and Ruslan Salakhutdinov.
\newblock Revisiting semi-supervised learning with graph embeddings.
\newblock In \emph{Proceedings of the 33rd International Conference on
  International Conference on Machine Learning - Volume 48}, ICML'16, pages
  40--48. JMLR.org, 2016.
\newblock URL \url{http://dl.acm.org/citation.cfm?id=3045390.3045396}.

\bibitem[You et~al.(2018{\natexlab{a}})You, Ying, Ren, Hamilton, and
  Leskovec]{graphrnncode}
Jiaxuan You, Rex Ying, Xiang Ren, William Hamilton, and Jure Leskovec.
\newblock Code for {G}raph{RNN}: Generating realistic graphs with deep
  auto-regressive model.
\newblock \url{https://github.com/JiaxuanYou/graph-generation},
  2018{\natexlab{a}}.

\bibitem[You et~al.(2018{\natexlab{b}})You, Ying, Ren, Hamilton, and
  Leskovec]{you18graphrnn}
Jiaxuan You, Rex Ying, Xiang Ren, William Hamilton, and Jure Leskovec.
\newblock {G}raph{RNN}: Generating realistic graphs with deep auto-regressive
  models.
\newblock In Jennifer Dy and Andreas Krause, editors, \emph{Proceedings of the
  35th International Conference on Machine Learning}, volume~80 of
  \emph{Proceedings of Machine Learning Research}, pages 5708--5717,
  Stockholmsmässan, Stockholm Sweden, 10--15 Jul 2018{\natexlab{b}}. PMLR.
\newblock URL \url{http://proceedings.mlr.press/v80/you18a.html}.

\bibitem[Zitnik and Leskovec(2017)]{Zitnik2017_ppidata}
Marinka Zitnik and Jure Leskovec.
\newblock Predicting multicellular function through multi-layer tissue
  networks.
\newblock \emph{Bioinformatics}, 33\penalty0 (14):\penalty0 190--198, 2017.

\end{thebibliography}
